\documentclass[11pt]{article}

\usepackage[preprint]{acl}

\usepackage{times}
\usepackage{latexsym}

\usepackage[T1]{fontenc}

\usepackage[utf8]{inputenc}

\usepackage{microtype}

\usepackage{inconsolata}

\usepackage{graphicx}

\usepackage{amssymb}
\usepackage{amsmath}
\usepackage{multirow}
\usepackage{pifont}

\usepackage{tcolorbox}
\tcbuselibrary{skins, breakable}

\newtcolorbox{promptbox}[1][]{
  colback=white,
  colframe=black,
  fonttitle=\bfseries,
  title=Prompt,
  boxrule=0.8pt,
  arc=3pt,
  breakable,
  #1
}

\newcommand{\red}[1]{\textcolor{red}{#1}}

\title{Reward Modeling for Scientific Writing Evaluation}

\author{Furkan \c{S}ahinu\c{c}$^{1,2}$, Subhabrata Dutta$^{1}$, Iryna Gurevych$^{1,2}$ \\
$^{1}$Ubiquitous Knowledge Processing Lab (UKP Lab) \\
Department of Computer Science and Hessian Center for AI (hessian.AI) \\
Technical University of Darmstadt \\
$^{2}$Konrad Zuse School of Excellence in Learning and Intelligent Systems (ELIZA) \\
\url{www.ukp.tu-darmstadt.de}}

\begin{document}
\maketitle

\begin{abstract}

Scientific writing is an expert-domain task that demands deep domain knowledge, task-specific requirements and reasoning capabilities that leverage the domain knowledge to satisfy the task specifications. While scientific text generation has been widely studied, its evaluation  remains a challenging and open problem. It is critical to develop models that can be reliably deployed for evaluating diverse open-ended scientific writing tasks while adhering to their distinct requirements. However, existing LLM-based judges and reward models are primarily optimized for general-purpose benchmarks with fixed scoring rubrics and evaluation criteria. Consequently, they often fail to reason over sparse knowledge of scientific domains when interpreting task-dependent and multi-faceted criteria. Moreover, fine-tuning for each individual task is costly and impractical for low-resource settings. To bridge these gaps, we propose cost-efficient, open-source reward models tailored for scientific writing evaluation. We introduce a two-stage training framework that initially optimizes scientific evaluation preferences and then refines reasoning capabilities. Our multi-aspect evaluation design and joint training across diverse tasks enable fine-grained assessment and robustness to dynamic criteria and scoring rubrics. Experimental analysis shows that our training regime strongly improves LLM-based scientific writing evaluation. Our models generalize effectively across tasks and to previously unseen scientific writing evaluation settings, allowing a single trained evaluator to be reused without task-specific retraining. We make our code\footnote{GitHub: \href{https://github.com/UKPLab/acl2026-expert-rm}{UKPLab/acl2026-expert-rm}} and data\footnote{Data: \href{https://tudatalib.ulb.tu-darmstadt.de/handle/tudatalib/4980}{TUdatalib}} publicly available.

\end{abstract}

\begin{figure}[h]
\centering
\includegraphics[width=0.916\columnwidth]{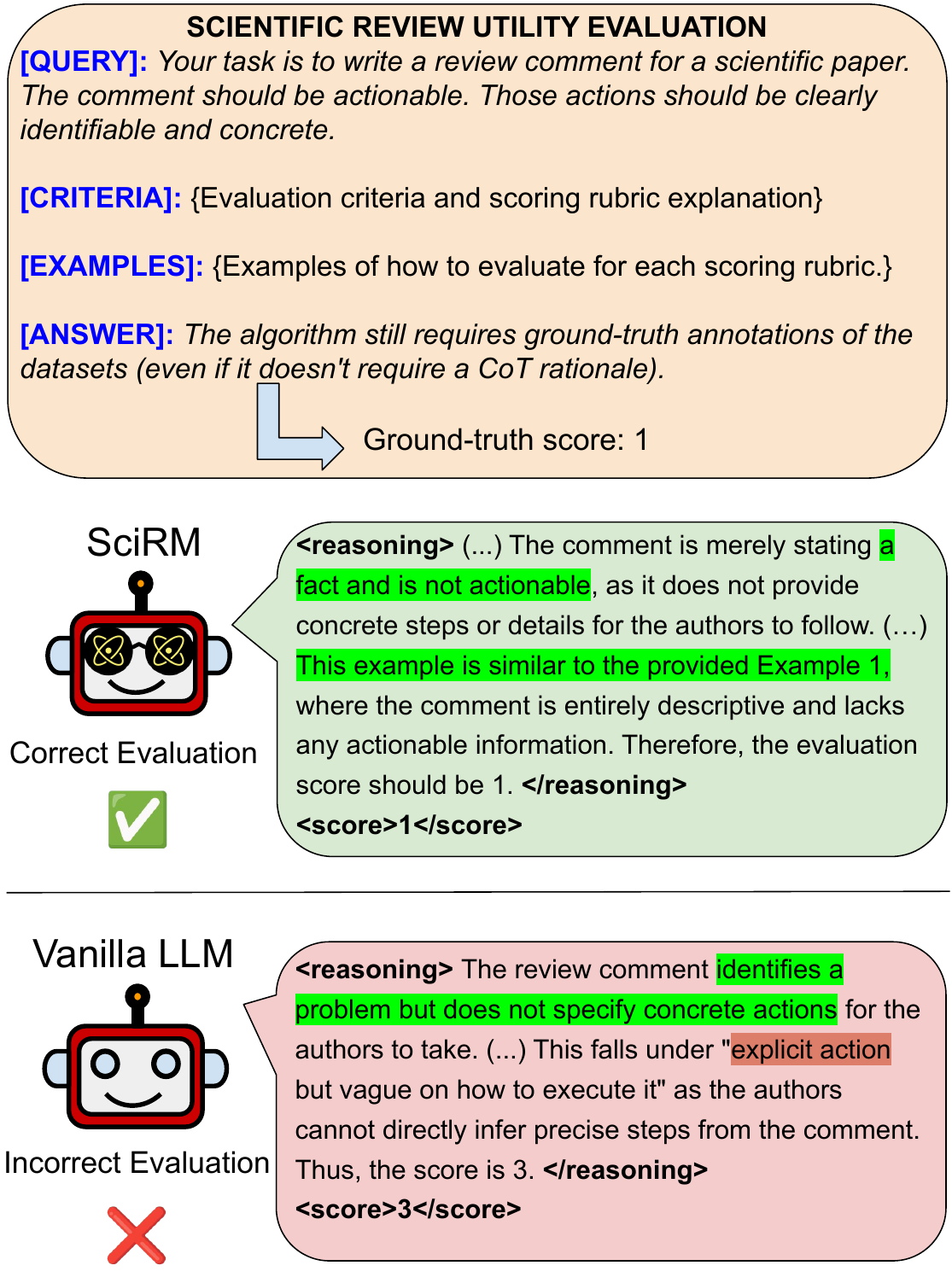}
\caption{Example demonstration of how we formalize scientific writing evaluation and the outputs of the review utility evaluation task. Vanilla LLM-based judges fail to properly reason over the task-specific evaluation criteria and provided examples. Contradictory statements are highlighted in different colors. In contrast, our \textsc{SciRM} model successfully incorporates the given criteria and examples into its reasoning process and correctly evaluates the scientific artifact.}
\label{fig:teaser}
\end{figure}

\section{Introduction}

\begin{figure*}[h]
\centering
\includegraphics[width=0.9\textwidth]{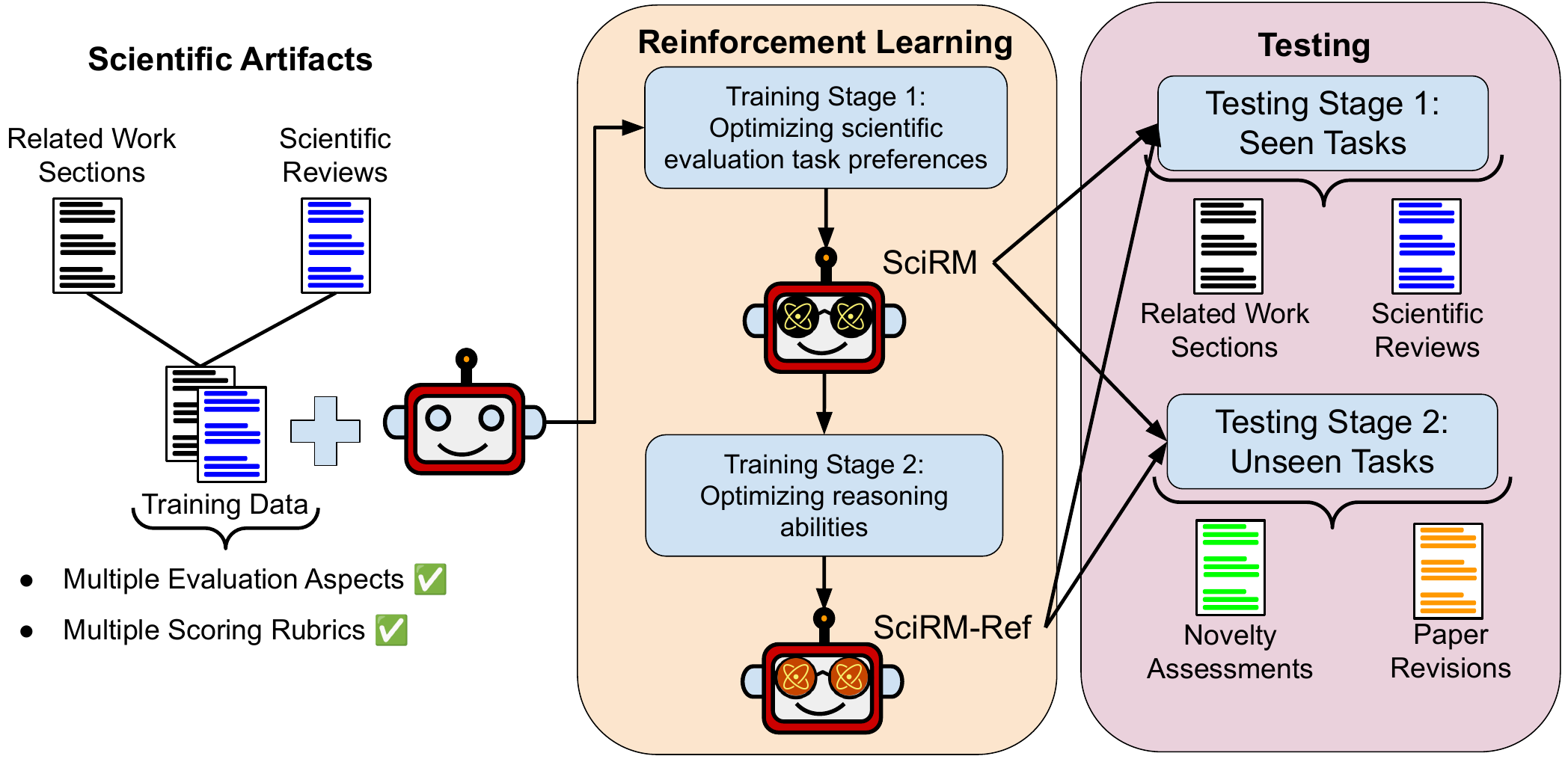}
\caption{Overview of \textsc{SciRM} and \textsc{SciRM-Ref} training and testing pipeline. Diverse scientific artifacts are used to construct training data with multiple evaluation aspects and scoring rubrics (see Section \ref{sec:dataset} for details). Models are trained via GRPO in two stages to optimize task specifications and reasoning capabilities, and are evaluated on both seen and unseen scientific writing evaluation tasks.}
\label{fig:framework}
\end{figure*}

Due to the strong text generation capabilities of large language models (LLMs), their application to scientific text generation, such as related work generation, review generation, and revising papers, has gained increasing attention recently \cite{Li:2024rw-survey,Liang:2024,Afzal:2026}. However, without proper evaluation, it is difficult to assess the accuracy and reliability of the generated texts. Therefore, the limitations in evaluation can block the entire development pipeline. Since scientific writing tasks have diverse and task-specific requirements, developing appropriate evaluation frameworks is a challenging problem. Training evaluators for each individual task is costly and, in some cases, infeasible due to limited data availability.

LLM-as-a-judge approaches \cite{Liu:2023,Zheng:2023} are the most widely adopted evaluation paradigms in scientific writing tasks. However, they often fail to reason over the given domain knowledge and task-specific preferences (see Figure \ref{fig:teaser}). This motivates the need for mechanisms that allow models to reason over and remain grounded in explicit evaluation guidelines (aka, \textit{constitution}) at inference time. Inference-time adaptability is a major challenge for existing approaches like Constitutional AI~\cite{bai2022constitutionalaiharmlessnessai}, which internalize fixed constitution during training and therefore cannot be readily applied to a diverse set of evaluation tasks. This rigidity is problematic while evaluating scientific text generations, as evaluation guidelines can be different, even contradictory, across different aspects, tasks, or domains.

To improve the reasoning capabilities of LLM-based judges, training reward reasoning models has recently gained popularity \cite{Ankner:2024,Guo:2025,Chen:2025judgelrm,Chen:2026,Wang:2024}. However, existing reward models are primarily optimized for improving performance on community-standard benchmarks, such as mathematical reasoning, instruction-following, and human-preference modeling for coding, helpfulness, and safety \cite{Lambert:2025,Malik:2026,Frick:2025}. Therefore, they fall short of capturing the nuanced requirements of scientific writing evaluation. In addition, the vast majority of reward models encode task preferences in a pairwise manner, which prevents independent assessment of text quality based on explicit task-specific criteria. Another drawback of such models is that they are optimized for fixed scoring rubrics and criteria. Scientific tasks have unique characteristics; each requires domain-specific expertise and evaluation dynamics that differ from those of open-ended creative writing tasks \cite{Chakrabarty:2025}. Even for the same scientific artifact, it is possible to evaluate from multiple aspects requiring different criteria and rubrics. However, current reward models experience performance degradation when they are applied to other tasks with evaluation rubrics that differ from those seen during training \cite{Yang:2024}. On the other hand, \citet{Sahinuc:2025} offers insights on multiple-aspect evaluation for expert-domain tasks. However, their approach is limited to a single task, and their most accurate pipeline relies on proprietary LLMs, which restricts scalability and hinders generalization across diverse scientific writing tasks.

In this work, we adapt reward model training strategies to enhance LLMs’ capabilities regarding \textit{what to evaluate} and \textit{how to evaluate} scientific writing tasks. Concretely, we design reward models that are conditioned on an explicit evaluation constitution—a structured description of criteria and label space—present during both training and inference. Furthermore, we introduce a two-stage optimization process in which models not only learn to follow the constitution but also reflexively reinterpret it to correct and stabilize their own reasoning. This process involves a joint optimization of in-context preference-following and reasoning abilities, which \cite{Lai:2024} identifies as a missing piece in modern LMs trained using RL. 

\textbf{Contributions and Findings:} We introduce cost-efficient reward models, \textsc{SciRM} and \textsc{SciRM-Ref}, specifically designed for scientific writing evaluation \textbf{(C1)}. We employ two-stage reinforcement learning to optimize the models for (1) scientific writing evaluation preferences and (2) reasoning abilities to better comprehend the given evaluation criteria, enabling models to explicitly reason over and faithfully adhere to dynamically specified evaluation rules. \textbf{(C2)}. Rather than producing a single aggregated score, our models evaluate scientific artifacts across multiple aspects. This approach enhances both reliability and interpretability of our evaluation \textbf{(C3)}. We curate and process datasets from diverse sources and jointly train our models across multiple tasks to (1) improve robustness against varying scoring rubrics and (2) enhance our models' generalization capabilities \textbf{(C4)}. We illustrate the overview of our pipeline in Figure \ref{fig:framework}. 

We finally test our models on four different scientific writing tasks: related work sections, paper reviews, novelty summary assessments, and paper revisions based on instructions, each with distinct evaluation aspects and scoring rubrics. Experimental results show that our two-stage training scheme substantially boosts the LLMs' scientific writing evaluation performance \textbf{(F1)}. In particular, our second training stage leads to improvements in tasks requiring strong reasoning capabilities \textbf{(F2)}. Furthermore, our models outperform baseline models on tasks that are not included in the training, indicating our model's strong generalization and scaling capabilities \textbf{(F3)}.

\section{Related Work}

\subsection{Scientific Writing Evaluation} 

Utilizing LLMs as direct evaluators is one of the most intuitive approaches for scientific writing evaluation due to their flexibility in being prompted with various task settings without additional training overhead \cite{Liu:2023,Zheng:2023}. However, prior work has shown that vanilla LLM-as-a-judge setups are prone to systematic biases and failures in domain-grounded reasoning \cite{Li:2024,Szymanski:2025,Gao:2025}. Moreover, the relative scarcity of scientific writing evaluation datasets also prevents LLMs from becoming more familiar with these tasks in the training phase.

To address these limitations, \citet{Jourdan:2025} focus on evaluating scientific text revision. They highlight that LLM-as-a-judge methods struggle to grasp task-specific evaluation aspects in the absence of a gold reference. In a complementary direction,  \citet{Purkayastha:2025} classify the peer-reviews according to common mistakes of reviewers referred to as \textit{lazy-thinking} patterns. A key limitation of their classification scheme is that, although some review sentences are suitable for multiple error categories, the dataset imposes a one-to-one label assignment. Similarly, \citet{Sadallah:2025} introduce a direct-evaluation benchmark that measures the utility of scientific reviews across the aspects of actionability, grounding, verifiability, and helpfulness.

Addressing a different task, \citet{Sahinuc:2025} propose a fine-grained evaluation framework for related work generation. Instead of producing a single overall score, they perform aspect based evaluation. Although their evaluation achieves strong alignment with human experts, their implementation is limited to the related work generation task.

\subsection{Evaluation-Tuned Models} 

Reward modeling has a significant impact on the success of post-training with reinforcement learning. Although verifiable rewards are computationally efficient and demonstrate strong performance in math and coding tasks \cite{Wen:2026}, many complex tasks are not suitable for directly verifiable reward signals, such as writing quality or helpfulness. These limitations and the recent success of reasoning-centric models on complex tasks \cite{Deepseekai:2025} have motivated the development of reward reasoning models \cite{Ankner:2024,Guo:2025,Chen:2025judgelrm,Chen:2026,Wang:2024}. The primary objective of such models is to generate more reliable and accurate rewards in non-verifiable tasks by leveraging intermediate thinking steps. Along with reasoning features, prior works also explore improving long-context comprehension \cite{Tang:2025}, integrating external documents \cite{Ma:2025}, and bridging pointwise and pairwise scoring paradigms in reward modeling settings \cite{Whitehouse:2026,Jian:2025}. The main drawback of these models is that they are optimized for standard reward benchmarks, which do not involve any scientific writing tasks. Therefore, they struggle to adapt to the varying scientific writing criteria and scoring. 

Although there are attempts to automatically generate task-specific evaluation criteria alongside the evaluation itself \cite{Liu:2025,Liu:2025dsgrm,Liang:2025}, the generated criteria are mostly surface-level and do not match the scientific writing evaluation requirements which are highly specific, diverse, and involve domain expertise.

Besides reward models, evaluation-tuned LLM-as-a-judge models also exist. In general, these models are not specialized for a particular task but are designed as general-purpose judge models \cite{Liu:2023,Kim:2024,Alexandru:2025,FlowJudge:2024,Skyworkcritic:2024}. This feature provides flexibility, allowing them to be employed in a diverse set of tasks. However, since they are not specialized for expert-domain tasks like scientific writing, they experience performance drops. In contrast, our work directly focuses on the evaluation of scientific writing and generalizes across different tasks with unique criteria and evaluation schemas.

\section{Methodology}

\subsection{Dataset} \label{sec:dataset}

To improve the generalization capabilities of our reward models, we include various scientific writing tasks that can be evaluated from multiple aspects using different scoring rubrics. These tasks focus on related work and review generation. \citet{Sahinuc:2025} provide a fine-grained set of aspects and principles for evaluating LLM-generated related work sections. From these aspects, we include coherence (whether citation sentences are consistent with the cited papers), positioning type (whether related works position themselves within existing literature, given a specific type), and positioning consistency (how well this positioning is maintained across different paragraphs) in our dataset. These aspects are subject to binary evaluation. To increase data diversity, we expand their dataset (originally containing NLP papers) by incorporating papers from other top-tier machine learning venues. We follow their preprocessing methodology to construct individual data instances. 

Additionally, we incorporate the RevUtil dataset \cite{Sadallah:2025}. The tasks in this dataset are evaluated on a 1–5 scale, with higher scores reflecting better quality. The number of instances and the corresponding scoring rubrics for each aspect of the training tasks are given in Appendix \ref{app:dataset}. In addition, example data instances can be found in evaluation prompts in Appendix \ref{app:prompts}.

\subsection{Reward Model Training}

Our training scheme consists of two distinct sequential stages: (1) teaching model to evaluate a candidate scientific text by reasoning upon an in-context constitution and (2) intensifying the reasoning acquired at stage 1 by teaching model self-verification based on the constitution. In both stages, we employ the GRPO algorithm due to its effectiveness in preference optimization and reasoning enhancement \cite{Shao:2024}. We formulate our task as follows: Given a prompt $x=(q,c,e)$, consisting of a task query $q$, criteria (or constitution) with a scoring rubric $c$ and example evaluations for each possible score $e$, generate an evaluation output $M(x)=y=(j,s)$ that consists of a reasoning (or justification) component $j$ along with a final score (or verdict) $s$.

To facilitate trivial extraction of rewards from the model outputs across different tasks and scoring rubrics, we use formatted outputs where the reasoning and scoring parts are encapsulated between \verb|<reasoning>| \verb|</reasoning>| and \verb|<score>| \verb|</score>| tags, respectively. The reasoning part helps to improve evaluation performance, as shown in previous works \cite{Kim:2025,Whitehouse:2026,Ma:2025}. We calculate our rewards based on compliance with the formatting and correctness of the generated score. In particular, let $y$ be the output, and let $m(y)$ be the score assigned by the model within \verb|<score>| \verb|</score>| in $y$, or $m(y)=\varnothing$ if no valid tag is present.
Let $s^*$ be the gold score and let $S$ be the set of possible scores in the corresponding scoring rubric. The reward function for the first training stage is defined as:
\begin{equation*}
r(y,s^*,S) =
\begin{cases}
-0.5 & \text{if } m(y)=\varnothing \\
0 & \text{if } m(y)\notin \mathbb{R} \\
0.25 & \text{if } m(y)\notin S \\
0.5 & \text{if } m(y)\in S \text{ \& } m(y)\neq s^*\\
1.5 & \text{if } m(y)\in S \text{ \& } m(y)=s^* \\
\end{cases}
\end{equation*}

This rewarding strategy helps identify different error types, including formatting and semantic errors. Our initial experiments showed that models may attempt to exploit the reward function \cite{Skalse:2022} by generating responses that only focus on the scoring criteria. We also observed occasional instability during training, where  models would fail to finish their outputs and continue generating until the maximum length was reached. To mitigate these issues, we introduce a length penalty function to mitigate training instability and reward hacking.
\begin{equation*}
f(L,T) =
\begin{cases}
\frac{(kT - L)^2}{kT^{\,2}}, & L < kT, \\[1em]
\frac{(L - (1-k)T)^2}{(T - (1-k)T)^{\,2}}, & L > (1-k)T, \\[1em]
0, & \text{otherwise}
\end{cases}
\end{equation*}

Here, $L$ and $T$ stand for the number of generated tokens and the maximum token length, respectively. The parameter $k \in (0, 0.5]$ controls how strict the penalty will be. When the generated output is too short or too long—i.e., closer to the boundary regions, the penalty quadratically increases with the distance from the acceptable range. If the length is within this range, no penalty is applied. The final reward is calculated by 
\begin{equation*}
R(x) = r(M(x),s^*,S) - f(L,T).
\end{equation*}

To further enhance the reasoning capabilities of our model, we introduce a second reinforcement learning stage, where the model is incentivized to update its evaluation by reflecting on the constitution. We first obtain the scores and reasoning from the model on the data instances. Next, we remove the scoring part and append the reasoning part with a reflection prompt that encourages the model to revisit the query and criteria before outputting its final score. The adjusted reward function for the second stage training is as follows:

\begin{equation*}
r_p(s_i,s_f, s^*) =
\begin{cases}
0.5 & \text{if } s_i = s^* \text{ \& } s_f = s^* \\
1.0 & \text{if } s_i \neq s^* \text{ \& } s_f = s^* \\
-0.5 & \text{if } s_i \neq s^* \text{ \& } s_f \neq s^* \\
-1.0 & \text{if } s_i = s^* \text{ \& } s_f \neq s^*
\end{cases}
\end{equation*}

$s_i$ and $s_f$ denote the initial and final scores, respectively. This rewarding strategy depends not only on the correctness of the final decision but also on the model's prior performance. If the model corrects itself, it receives the highest reward; when it backslides, it is harshly penalized. Consistent behaviors yield a moderate reward or penalty based on correctness. In other words, we promote the model to revisit the criteria and its earlier reasoning either to fix the incorrect trajectory or to maintain the correct one. We refer to the model trained only in the first stage as \textsc{SciRM}, and the model trained using the two-stage procedure as \textsc{SciRM-Ref}.

\section{Experiments}

\textbf{Baselines:} We use Qwen 2.5-7B \cite{Qwen2.5:2025} as our base model, based on its performance in our preliminary experiments on scientific writing evaluation. We evaluate our model against several open-weight LLMs of comparable size, representing different model families: DS-R1-Qwen-7B \cite{Deepseekai:2025}, 
Granite3.3-8B-Instruct \cite{IBM:2025},
Llama3.1-8B-Instruct \cite{Grattafiori:2024},
Qwen2.5-7B-Instruct \cite{Qwen2.5:2025}, and
Qwen3-8B \cite{Yang:2025}. In addition to open-weight models, we experiment with GPT-5.2 (2025-12-11) and o3-mini (2025-01-31), which was used as a backbone model in related work evaluation by \citet{Sahinuc:2025}, to gain insight about the performance bounds of large proprietary models. 

We also include specialized evaluation models in our baseline set. To represent tuned, general purpose LLM-as-a-judge models, we use Prometheus\footnote{\href{https://huggingface.co/Unbabel/M-Prometheus-7B}{M-Prometheus-7B}} \cite{Kim:2024} and Selene\footnote{\href{https://huggingface.co/AtlaAI/Selene-1-Mini-Llama-3.1-8B}{Selene-1-Mini-Llama-3.1-8B}} \cite{Alexandru:2025} frameworks. Furthermore, based on open availability and compatibility with pointwise scoring, we include DeepSeek-GRM\footnote{\href{https://huggingface.co/BBQGOD/DeepSeek-GRM-16B}{DeepSeek-GRM-16B}} \cite{Liu:2025dsgrm} and Skywork-Critic\footnote{\href{https://huggingface.co/Skywork/Skywork-Critic-Llama-3.1-70B}{Skywork-Critic-Llama-3.1-70B }} \cite{Skyworkcritic:2024} as reward modeling baselines.

\textbf{Experimental Setting:} We employ LoRA framework \cite{Hu:2022} to meet the memory requirements and facilitate training. Besides testing our reward models on evaluation of related works and scientific reviews, we introduce two additional experimental settings. For each experiment, the inference is repeated five times (GPT-5.2 and o3-mini three times) on test sets with a temperature of 1 and a top-p of 0.95, and we report the resulting mean and standard deviation. Further experimental details and evaluation prompts are provided in Appendix \ref{app:exp} and \ref{app:prompts} respectively.

\subsection{Unseen Aspect Experiments} 

To evaluate whether our training scheme provides an improvement over the baselines on unseen aspects of the same evaluation task, we conduct an experiment that excludes selected evaluation aspects from the training data. Specifically, we remove the \textit{Actionability} and \textit{Grounding} aspects from the RevUtil dataset and test whether models trained on the remaining aspects can still improve performance on unseen evaluation aspects. We refer to the resulting model as \textsc{SciRM-Masked}. 

\subsection{Unseen Task Experiments} To assess the generalization capabilities of our reward model, we test it on two additional tasks that are not included in the training data: comparison of novelty assessments and evaluation scientific revision quality. For the first task, we utilize the novelty-evaluation dataset from \cite{Afzal:2026}, which contains summaries of both human and LLM-generated novelty assessments. Human summaries are accompanied by human-annotated binary labels indicating the novelty of the papers. However, LLM-generated assessment summaries do not include final novelty labels. To address this, we use GPT-5.1 \cite{Openai:2025} to infer the novelty labels from the LLM-generated summaries. Our evaluation objective is to assess whether LLM-generated summaries align with human summaries in terms of their novelty verdicts (i.e., \textit{novel}, \textit{not novel}). We treat equality of the novelty labels as the ground-truth indicator of alignment between summary pairs. In the final evaluation dataset, there are 76 assessment pairs with 56 aligned pairs and 20 non-aligned pairs.

For the second task, we use the scientific revision dataset from \cite{Jourdan:2025}, which includes paragraphs from scientific papers and corresponding LLM-generated revisions for a given instruction. Each candidate revision is annotated along two dimensions: whether it addresses the given instruction (i.e., \textit{relatedness}) and whether it correctly applies the instruction to produce an improved version of the original text (i.e., \textit{correctness}). For each aspect, we have 3,092 data instances, resulting in a total of 6,184 data points.

\section{Results}

\begin{figure*}[h]
\centering
\includegraphics[width=1.0\textwidth]{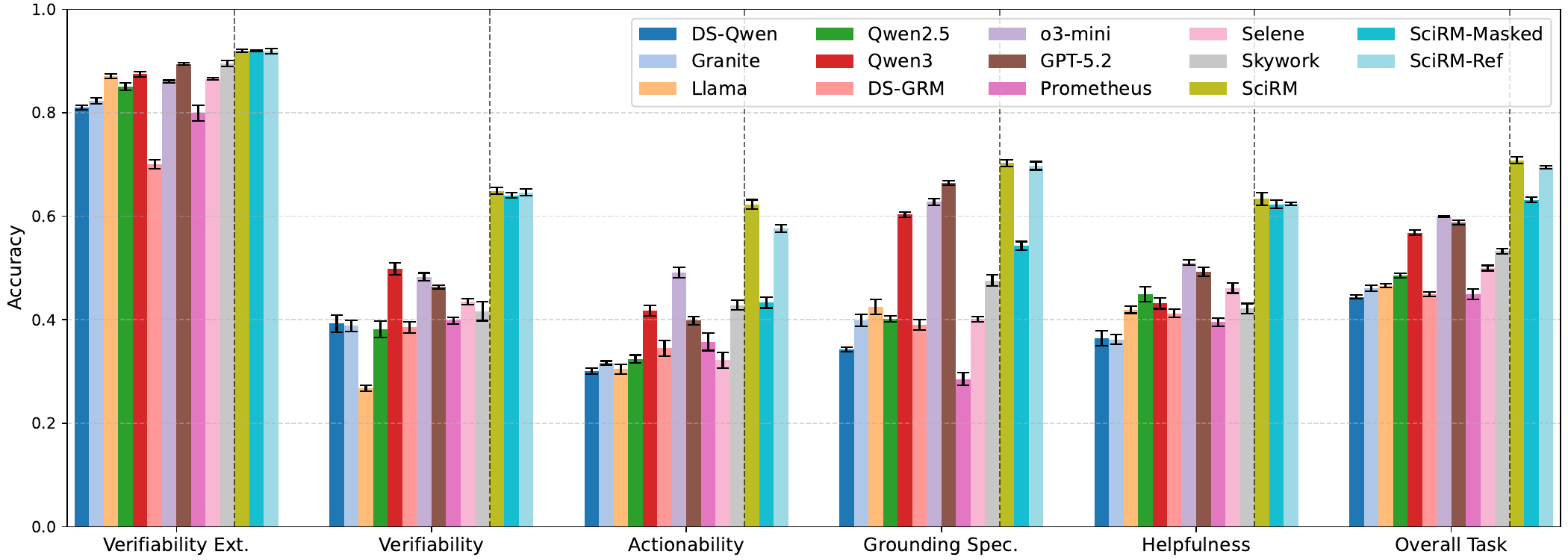}
\caption{Review utility evaluation results. Results to the left of the vertical dashed lines correspond to the baseline models, whereas results to the right correspond to our models.}
\label{fig:rev_util}
\end{figure*}

\begin{figure*}[h]
\centering
\includegraphics[width=1.0\textwidth]{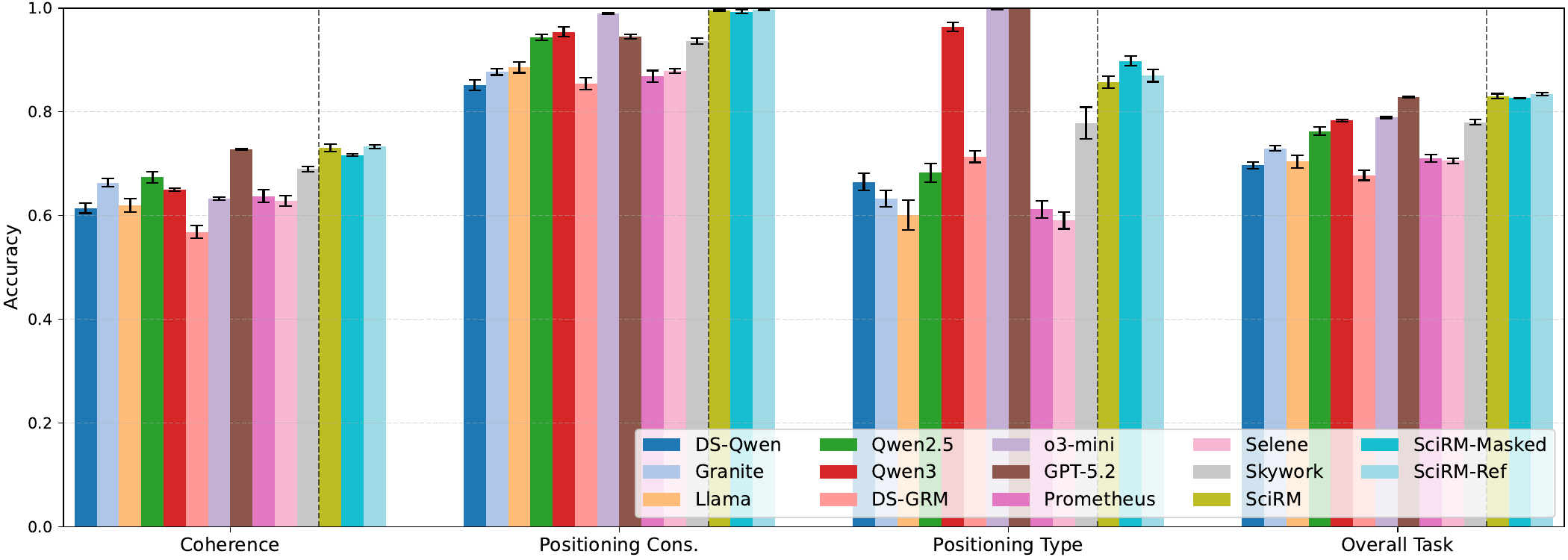}
\caption{Related work evaluation results. Results to the left of the vertical dashed lines correspond to the baseline models, whereas results to the right correspond to our models.}
\label{fig:rw_gen}
\end{figure*}

\textbf{Performance on seen domains.} We begin by evaluating our models on the tasks included in the training phase. For review utility evaluation, we present results in Figure \ref{fig:rev_util}. We observe that our training scheme leads to a clear improvement over baseline models. Both \textsc{SciRM} and \textsc{SciRM-Ref} consistently achieve the highest accuracy across all evaluation aspects. Among these aspects, all models reach their highest scores on \textit{Verifiability Extraction}. Since this aspect focuses on binary claim detection in reviews rather than scoring on a 1–5 scale, it constitutes a comparatively simpler evaluation than the other aspects, which require a deeper understanding of the evaluation criteria. Across the remaining aspects, our models perform comparably among themselves, which is also observed for the baseline models.

We report the model performances for the related work evaluation in Figure \ref{fig:rw_gen}. We observe a similar pattern to the results in the review utility evaluation in terms of improvements over the baseline models. In particular, our models achieve near-perfect accuracy on \textit{Positioning Consistency} aspect. In contrast, \textit{Coherence} stands out as a more challenging aspect, since comparing citation sentences with the cited papers requires a higher degree of semantic understanding. Although the performance gap between our models and the baselines is largest for \textit{Positioning Type}, Qwen3, GPT-5.2, and o3-mini exhibit outlier behavior on this aspect, possibly due to better leveraging their reasoning capabilities. However, our models still outperform all baselines when all data instances are considered.

Figures \ref{fig:rev_util} and \ref{fig:rw_gen} collectively indicate that our training regime helps to leverage the comprehension capabilities of LLM models in scientific writing evaluation. Furthermore, our joint training across multiple tasks improves the models' adaptability to diverse scoring rubrics and evaluation criteria. For completeness, we report full numerical results, including mean and standard deviation values for accuracy and F1 score in Appendix \ref{app:full_results}. 

\noindent \textbf{Unseen Aspect Performance.} We further evaluate performance on unseen aspects by reporting the results of \textsc{SciRM-Masked} in Figure \ref{fig:rev_util}. Since the \textit{Actionability} and \textit{Grounding} aspects are excluded from the training, the performance on these aspects decreases compared to our other models. Nevertheless, \textsc{SciRM-Masked} continues to outperform the baseline models, with the exception of Qwen3, GPT-5.2, and o3-mini on the \textit{Grounding} aspect. This behavior suggests that our methodology enables the model to capture the underlying structure of the task, rather than overfitting to specific evaluation aspects observed during training. As a result, the model can generalize the learned reasoning patterns and adapt effectively to previously unseen evaluation criteria, even when explicit supervision for those aspects is absent. For the other aspects included in training for both revision utility and related work, \textsc{SciRM-Masked} performs similar to our other model variants. 

\begin{table}[!t]
\centering
\begin{tabular}{l|c|ccc}
\hline
\multirow{2}{*}{\textbf{Model}} & \textbf{Novelty} & \multicolumn{3}{c}{\textbf{Revision}} \\
& \textbf{Align.} & \textbf{Rel.} & \textbf{Cor.} & \textbf{Overall} \\
\hline
DS-Qwen & 0.41 & 0.77 & 0.67 & 0.72 \\
Granite3.3 & 0.48 & 0.80 & 0.68 & 0.74 \\
Llama3.1 & 0.56 & 0.56 & 0.61 & 0.58 \\
Qwen2.5 & 0.71 & 0.80 & 0.66 & 0.73 \\
Qwen3 & 0.58 & 0.86 & 0.76 & 0.81 \\
\hline
Prometheus & 0.33 & 0.76 & 0.70 & 0.73 \\
Selene & 0.57 & 0.79 & 0.71 & 0.75 \\
Skywork & 0.42 & 0.83 & 0.77 & 0.80 \\
DS-GRM & 0.35 & 0.73 & 0.59 & 0.66 \\
\hline
o3-mini & 0.78 & 0.88 & 0.80 & 0.84 \\
GPT-5.2 & 0.74 & 0.82 & 0.78 & 0.80 \\
\hline
SciRM & 0.61 & 0.82 & 0.73 & 0.78 \\
SciRM-Ref & 0.74 & 0.82 & 0.73 & 0.78 \\
\hline
\end{tabular}
\caption{Model performance in terms of accuracy for unseen tasks. Rel. and Cor. denotes Relatedness and Correctness aspects for revision evaluation task.}
\label{tab:unseen}
\end{table}

\noindent \textbf{Unseen Task Performance.} Table \ref{tab:unseen} shows the performance of our models on tasks that are used exclusively as test sets. In the novelty alignment evaluation, \textsc{SciRM-Ref} not only outperforms the baseline models but also shows a clear improvement over \textsc{SciRM}. Among all models, o3-mini achieves the highest accuracy, while \textsc{SciRM-Ref} delivers near–closed-model performance despite having smaller size and being fully open. We note that novelty evaluation differs from other tasks, as determining the novelty of a paper based on summarized reviewer comments and comparing these according to their implicit novelty judgments requires strong reasoning capabilities. These results indicate that the second training stage is effective in further enhancing the model’s reasoning ability beyond what is achieved in the initial training stage. Although we provide explicit criteria to illustrate how the evaluation should be performed, only \textsc{SciRM-Ref} is able to effectively leverage this additional supervision.

In the paper revision evaluation, our models either outperform baselines or perform comparably to the proprietary and larger models. We observe that models generally tend to perform better on the \textit{Relatedness} aspect than on \textit{Correctness}. We note that this task primarily evaluates instruction-following on scientific text. Since it involves instructions like \textit{content addition or removal, summarization, and merging}, implementing evaluations is relatively less reasoning-intensive than in novelty alignment. 

Overall, the observed performance gains indicate improved cross-task generalization rather than memorization. Since the unseen tasks are diverse and substantially different from those in training, consistent improvements under this distribution shift suggest that the model has learned transferable capabilities instead of relying on task-specific patterns. These capabilities enable the model to apply learned evaluation principles across diverse scientific writing scenarios.

\begin{table}[!t]
\centering
\renewcommand{\arraystretch}{1.0}
\begin{tabular}{l|cccc|c}
\hline
\textbf{Model} & \textbf{RW} & \textbf{RU} & \textbf{Nov.} & \textbf{Rev.} & \textbf{Avg.} \\
\hline
DS-Qwen & 0.70 & 0.44 & 0.41 & 0.72 & 0.57 \\
Granite3.3 & 0.73 & 0.46 & 0.48 & 0.74 & 0.60 \\
Llama3.1 & 0.70 & 0.47 & 0.56 & 0.58 & 0.58 \\
Qwen2.5 & 0.76 & 0.49 & 0.71 & 0.73 & 0.67 \\
Qwen3 & 0.78 & 0.57 & 0.58 & 0.81 & 0.68 \\
\hline
Prometheus & 0.71 & 0.45 & 0.33 & 0.73 & 0.56 \\
Selene & 0.71 & 0.50 & 0.57 & 0.75 & 0.63 \\
Skywork & 0.78 & 0.53 & 0.42 & 0.80 & 0.63 \\
DS-GRM & 0.68 & 0.45 & 0.35 & 0.66 & 0.54 \\
\hline
o3-mini & 0.79 & 0.60 & {0.78} & {0.84} & 0.75 \\
GPT-5.2 & {0.83} & 0.59 & 0.74 & 0.80 & 0.74 \\
\hline
SciRM & {0.83} & {0.71} & 0.61 & 0.78 & 0.73 \\
SciRM-Ref & {0.83} & 0.69 & 0.74 & 0.78 & {0.76} \\
\hline
\end{tabular}
\caption{Individual and average task performances. (RW: Related work, RU: Review utility, Nov: Novelty alignment, Rev: Revision alignment, Avg: Average)}
\label{tab:avg}
\end{table}

In Table \ref{tab:avg}, we summarize the results for all four evaluation tasks. Our reward models outperform the baselines on individual tasks and result in a larger average score. \textsc{SciRM-Ref} becomes the prominent model by achieving the largest average score. Although it exhibits very similar results to \textsc{SciRM} on the related work, review utility, and revision evaluation tasks, the difference arises from the novelty alignment evaluation task, in which the reasoning capabilities of the models are crucial.

\begin{table}[t]
\centering
\begin{tabular}{l|c|c}
\hline
\textbf{Models} & \textbf{Reasoning Difference} & \textbf{Example} \\
\hline
SciRM & Grounded reasoning vs & \multirow{2}{*}{Table \ref{tab:scirm-qwen2.5}} \\
Qwen2.5 & speculative inference & \\
\hline
SciRM & Rubric-alignment vs & \multirow{2}{*}{Table \ref{tab:scirmref-scirm}} \\
SciRM-Ref & semantic misweighting & \\
\hline
SciRM-Ref  & Detail sensitivity vs & \multirow{2}{*}{Table \ref{tab:scirmref-dsgrm}} \\
DS-GRM & content omission & \\
\hline
\end{tabular}
\caption{Reasoning trace comparison summary table. Corresponding full trace tables are provided in Example column.}
\label{tab:reasoning}
\end{table}

\section{Model Analysis and Ablation Studies}

\subsection{Reasoning Traces and Error Analysis.}

To obtain further insight into the mechanism behind the improvements of our models, we make a comprehensive analysis on generated reasoning traces across various models and tasks. First, we observe that baseline models tend to make assumptions beyond the provided evaluation criteria rather than fully grounding their judgments in them, as our model does (e.g., inferring implicit actionability despite the absence of explicit actionable guidance.).

Secondly, our second-stage training improves rubric alignment in evaluation, particularly for the novelty alignment task. Without reasoning-focused training, models may overemphasize certain aspects of the assessments (e.g., their characterization as incremental contributions) while disregarding other explicit novelty statements, leading to incorrect conclusions about alignment.

Finally, baseline models can make more direct errors, such as failing to fully comprehend the evaluated text, for example by overlooking the changed text in the scientific revision task. Furthermore, our manual analysis shows that the reasoning traces generated by our models are fully consistent with their final verdicts. Table \ref{tab:reasoning} summarizes the qualitative differences in generated reasoning and provides references to the full reasoning traces in Appendix \ref{app:reasoning}.

\subsection{Ablation Study}

To investigate the role of the reasoning to guide the final judgment, we ablate the reasoning process for our proposed models. Specifically, we remove the instructions for reasoning process in the system prompt and example reasoning parts in evaluation constitution. In Table \ref{tab:ablation}, we observe that both individual and overall average task performance decrease in the absence of reasoning traces. This suggests that reasoning-conditioned final judgments lead to more accurate results. Importantly, novelty alignment evaluation suffers most from the ablation for both models, as it requires non-trivial generalization of reasoning and grounding into the constitution. This further strengthens our claim that our method introduces generalization to unseen tasks.   

\begin{table}[!t]
\centering
\renewcommand{\arraystretch}{1.0}
\resizebox{\columnwidth}{!}{
\begin{tabular}{l|cccc|c}
\hline
\textbf{Model} & \textbf{RW} & \textbf{RU} & \textbf{Nov.} & \textbf{Rev.} & \textbf{Avg.} \\
\hline
SciRM & \red{-0.03} & \red{-0.06} & \red{-0.21} & \red{-0.00} & \red{-0.07} \\
SciRM-Ref & \red{-0.02} & \red{-0.05} & \red{-0.25} & \red{-0.01} & \red{-0.08} \\
\hline
\end{tabular}
}
\caption{Ablation study on the reasoning process across tasks. Numbers denote the performance change relative to the full pipeline (Table \ref{tab:avg}) when reasoning is disabled at evaluation time.}
\label{tab:ablation}
\end{table}

\section{Conclusion}

In this paper, we address the problem of evaluating scientific writing tasks. Owing to their unique characteristics, both the generation and evaluation of scientific writing require greater attention than common, well-defined NLP tasks. Moreover, this uniqueness often results in limited annotated data, making it impractical to build specialized models for individual tasks. To address this challenge, we propose reward models that enhance reasoning capabilities and capture task dependencies through explicit evaluation criteria and scoring rubrics. Our models outperform strong baselines in both seen and unseen task settings, highlighting their reasoning and generalization capabilities and underscoring their practicality for real-world scientific workflows. This design enables a single trained evaluator to generalize across tasks, substantially reducing the need for task-specific retraining.

We believe that this work takes a step toward more reliable, interpretable, and reusable evaluation frameworks for scientific writing. A promising direction for future work is to improve scientific text generation across diverse tasks using reinforcement learning, where rewards are provided by our models. By producing fine-grained, aspect-level feedback on scientific text quality, our models can guide generators along multiple evaluation dimensions rather than optimizing a single objective. Finally, the cost efficiency of our open-source models further enhances their practicality compared to proprietary LLM-based evaluators.

\section*{Limitations}

Due to the computational requirements of GRPO, we restrict our experiments to 7B-scale models. We hypothesize that larger models could better capture task-specific evaluation nuances, generalize more effectively, and exhibit more accurate reasoning. Nevertheless, our training scheme provides an opportunity to develop small-sized evaluation models that can outperform or remain competitive with proprietary models. Another limitation is the scarcity of evaluation datasets for scientific writing. This results in most of the scoring rubrics being binary. With the introduction of more fine-grained scientific writing evaluation datasets, a more comprehensive analysis can be conducted as future work.

\section*{Acknowledgments}

This work has been funded/co-funded by the European Union (ERC, InterText, 101054961). Views and opinions expressed are however those of the author(s) only and do not necessarily reflect those of the European Union or the European Research Council. Neither the European Union nor the granting authority can be held responsible for them. Furkan \c{S}ahinu\c{c} is supported by the Konrad Zuse School of Excellence in Learning and Intelligent Systems (\href{https://eliza.school/}{ELIZA}) through the DAAD programme Konrad Zuse Schools of Excellence in Artificial Intelligence, sponsored by the Federal Ministry of Education and Research. We also thank Sheng Lu, Tim Baumgärtner (UKP Lab, TU Darmstadt), and Huiyin Xue (University of Sheffield) for their feedback during the writing of this paper.

\bibliography{custom_v2}

@inproceedings{Sadallah:2025,
    title       = {The Good, the Bad and the Constructive: Automatically Measuring Peer Review{'}s Utility for Authors},
    author      = {Sadallah, Abdelrahman and Baumg{\"a}rtner, Tim and Gurevych, Iryna and Briscoe, Ted},
    booktitle   = {Proceedings of the 2025 Conference on Empirical Methods in Natural Language Processing},
    month       = nov,
    year        = {2025},
    address     = {Suzhou, China},
    publisher   = {Association for Computational Linguistics},
    url         = {https://aclanthology.org/2025.emnlp-main.1476/},
    doi         = {10.18653/v1/2025.emnlp-main.1476},
    pages       = {28979--29009},
}

@misc{Sahinuc:2025,
    title       = {Expert Preference-based Evaluation of Automated Related Work Generation}, 
    author      = {Furkan \c{S}ahinu\c{c} and Subhabrata Dutta and Iryna Gurevych},
    year        = {2025},
    eprint      = {2508.07955},
    archivePrefix={arXiv},
    primaryClass={cs.CL},
    url         = {https://arxiv.org/abs/2508.07955}, 
}

@misc{bai2022constitutionalaiharmlessnessai,
      title={Constitutional AI: Harmlessness from AI Feedback}, 
      author={Yuntao Bai and Saurav Kadavath and Sandipan Kundu and Amanda Askell and Jackson Kernion and Andy Jones and Anna Chen and Anna Goldie and Azalia Mirhoseini and Cameron McKinnon and Carol Chen and Catherine Olsson and Christopher Olah and Danny Hernandez and Dawn Drain and Deep Ganguli and Dustin Li and Eli Tran-Johnson and Ethan Perez and Jamie Kerr and Jared Mueller and Jeffrey Ladish and Joshua Landau and Kamal Ndousse and Kamile Lukosuite and Liane Lovitt and Michael Sellitto and Nelson Elhage and Nicholas Schiefer and Noemi Mercado and Nova DasSarma and Robert Lasenby and Robin Larson and Sam Ringer and Scott Johnston and Shauna Kravec and Sheer El Showk and Stanislav Fort and Tamera Lanham and Timothy Telleen-Lawton and Tom Conerly and Tom Henighan and Tristan Hume and Samuel R. Bowman and Zac Hatfield-Dodds and Ben Mann and Dario Amodei and Nicholas Joseph and Sam McCandlish and Tom Brown and Jared Kaplan},
      year={2022},
      eprint={2212.08073},
      archivePrefix={arXiv},
      primaryClass={cs.CL},
      url={https://arxiv.org/abs/2212.08073}, 
}

@misc{Shao:2024,
    title       = {DeepSeekMath: Pushing the Limits of Mathematical Reasoning in Open Language Models}, 
    author      = {Zhihong Shao and Peiyi Wang and Qihao Zhu and Runxin Xu and Junxiao Song and Xiao Bi and Haowei Zhang and Mingchuan Zhang and Y. K. Li and Y. Wu and Daya Guo},
    year        = {2024},
    eprint      = {2402.03300},
    archivePrefix={arXiv},
    primaryClass={cs.CL},
    url         = {https://arxiv.org/abs/2402.03300}, 
}

@misc{Kim:2025,
    title       = {Scaling Evaluation-time Compute with Reasoning Models as Process Evaluators}, 
    author      = {Seungone Kim and Ian Wu and Jinu Lee and Xiang Yue and Seongyun Lee and Mingyeong Moon and Kiril Gashteovski and Carolin Lawrence and Julia Hockenmaier and Graham Neubig and Sean Welleck},
    year        = {2025},
    eprint      = {2503.19877},
    archivePrefix={arXiv},
    primaryClass={cs.CL},
    url         = {https://arxiv.org/abs/2503.19877}, 
}

@inproceedings{Whitehouse:2026,
    title       = {J1: Incentivizing Thinking in {LLM}-as-a-Judge via Reinforcement Learning},
    author      = {Chenxi Whitehouse and Tianlu Wang and Ping Yu and Xian Li and Jason E Weston and Ilia Kulikov and Swarnadeep Saha},
    booktitle   = {Fourteenth International Conference on Learning Representations},
    year        = {2026},
    url         = {https://openreview.net/forum?id=dnJEHl6DI1}
}

@misc{Ma:2025,
    title       = {From Faithfulness to Correctness: Generative Reward Models that Think Critically}, 
    author      = {Qiyao Ma and Yunsheng Shi and Hongtao Tian and Chao Wang and Weiming Chang and Ting Yao},
    year        = {2025},
    eprint      = {2509.25409},
    archivePrefix={arXiv},
    primaryClass={cs.CL},
    url         = {https://arxiv.org/abs/2509.25409}, 
}

@inproceedings{Skalse:2022,
    author      = {Skalse, Joar and Howe, Nikolaus and Krasheninnikov, Dmitrii and Krueger, David},
    booktitle   = {Advances in Neural Information Processing Systems},
    pages       = {9460--9471},
    publisher   = {Curran Associates, Inc.},
    title       = {Defining and Characterizing Reward Gaming},
    url         = {https://proceedings.neurips.cc/paper_files/paper/2022/file/3d719fee332caa23d5038b8a90e81796-Paper-Conference.pdf},
    volume      = {35},
    year        = {2022}
}

@misc{Qwen2.5:2025,
    title       = {Qwen2.5 Technical Report}, 
    author      = {An Yang and Baosong Yang and Beichen Zhang and Binyuan Hui and Bo Zheng and Bowen Yu and Chengyuan Li and Dayiheng Liu and Fei Huang and Haoran Wei and Huan Lin and Jian Yang and Jianhong Tu and Jianwei Zhang and Jianxin Yang and Jiaxi Yang and Jingren Zhou and Junyang Lin and Kai Dang and Keming Lu and Keqin Bao and Kexin Yang and Le Yu and Mei Li and Mingfeng Xue and Pei Zhang and Qin Zhu and Rui Men and Runji Lin and Tianhao Li and Tianyi Tang and Tingyu Xia and Xingzhang Ren and Xuancheng Ren and Yang Fan and Yang Su and Yichang Zhang and Yu Wan and Yuqiong Liu and Zeyu Cui and Zhenru Zhang and Zihan Qiu},
    year        = {2025},
    eprint      = {2412.15115},
    archivePrefix={arXiv},
    primaryClass={cs.CL},
    url         = {https://arxiv.org/abs/2412.15115}, 
}

@misc{Unsloth:2023,
    author      = {Daniel Han, Michael Han, Unsloth team},
    title       = {Unsloth},
    url         = {http://github.com/unslothai/unsloth},
    year        = {2023}
}

@inproceedings{Hu:2022,
    title       = {Lo{RA}: Low-Rank Adaptation of Large Language Models},
    author      = {Edward J Hu and yelong shen and Phillip Wallis and Zeyuan Allen-Zhu and Yuanzhi Li and Shean Wang and Lu Wang and Weizhu Chen},
    booktitle   = {International Conference on Learning Representations},
    year        = {2022},
    url         = {https://openreview.net/forum?id=nZeVKeeFYf9}
}

@inproceedings{Lambert:2025,
    title       = {{R}eward{B}ench: Evaluating Reward Models for Language Modeling},
    author      = {Lambert, Nathan and Pyatkin, Valentina and Morrison, Jacob and Miranda, LJ and Lin, Bill Yuchen and Chandu, Khyathi and Dziri, Nouha and Kumar, Sachin and Zick, Tom and Choi, Yejin and Smith, Noah A. and Hajishirzi, Hannaneh},
    booktitle   = {Findings of the Association for Computational Linguistics: NAACL 2025},
    month       = apr,
    year        = {2025},
    address     = {Albuquerque, New Mexico},
    publisher   = {Association for Computational Linguistics},
    url         = {https://aclanthology.org/2025.findings-naacl.96/},
    doi         = {10.18653/v1/2025.findings-naacl.96},
    pages       = {1755--1797},
}

@inproceedings{Malik:2026,
    title       = {RewardEval: Advancing Reward Model Evaluation},
    author      = {Saumya Malik and Valentina Pyatkin and Sander Land and Jacob Morrison and Noah A. Smith and Hannaneh Hajishirzi and Nathan Lambert},
    booktitle   = {Fourteenth International Conference on Learning Representations},
    year        = {2026},
    url         = {https://openreview.net/forum?id=fb0G86Dewb}
}

@inproceedings{Frick:2025,
    title       = {How to Evaluate Reward Models for {RLHF}},
    author      = {Evan Frick and Tianle Li and Connor Chen and Wei-Lin Chiang and Anastasios Nikolas Angelopoulos and Jiantao Jiao and Banghua Zhu and Joseph E. Gonzalez and Ion Stoica},
    booktitle   = {Thirteenth International Conference on Learning Representations},
    year        = {2025},
    url         = {https://openreview.net/forum?id=cbttLtO94Q}
}

@inproceedings{Purkayastha:2025,
    title       = {{L}azy{R}eview: A Dataset for Uncovering Lazy Thinking in {NLP} Peer Reviews},
    author      = {Purkayastha, Sukannya and Li, Zhuang and Lauscher, Anne and Qu, Lizhen and Gurevych, Iryna},
    booktitle   = {Proceedings of the 63rd Annual Meeting of the Association for Computational Linguistics (Volume 1: Long Papers)},
    month       = jul,
    year        = {2025},
    address     = {Vienna, Austria},
    publisher   = {Association for Computational Linguistics},
    url         = {https://aclanthology.org/2025.acl-long.165/},
    doi         = {10.18653/v1/2025.acl-long.165},
    pages       = {3280--3308},
}

@inproceedings{Szymanski:2025,
    author      = {Szymanski, Annalisa and Ziems, Noah and Eicher-Miller, Heather A. and Li, Toby Jia-Jun and Jiang, Meng and Metoyer, Ronald A.},
    title       = {Limitations of the LLM-as-a-Judge Approach for Evaluating LLM Outputs in Expert Knowledge Tasks},
    year        = {2025},
    publisher   = {Association for Computing Machinery},
    address     = {New York, NY, USA},
    url         = {https://doi.org/10.1145/3708359.3712091},
    doi         = {10.1145/3708359.3712091},
    booktitle   = {Proceedings of the 30th International Conference on Intelligent User Interfaces},
    pages       = {952–966},
    series      = {IUI '25}
}

@article{Gao:2025,
    title       = {{LLM}-based {NLG} Evaluation: Current Status and Challenges},
    author      = {Gao, Mingqi and Hu, Xinyu and Yin, Xunjian and Ruan, Jie and Pu, Xiao and Wan, Xiaojun},
    journal     = {Computational Linguistics},
    volume      = {51},
    month       = jun,
    year        = {2025},
    address     = {Cambridge, MA},
    publisher   = {MIT Press},
    url         = {https://aclanthology.org/2025.cl-2.9/},
    doi         = {10.1162/coli_a_00561},
    pages       = {661--687},
}

@inproceedings{Li:2024,
    title       = {Leveraging Large Language Models for {NLG} Evaluation: Advances and Challenges},
    author      = {Li, Zhen and Xu, Xiaohan and Shen, Tao and Xu, Can and Gu, Jia-Chen and Lai, Yuxuan and Tao, Chongyang and Ma, Shuai},
    editor      = {Al-Onaizan, Yaser and Bansal, Mohit and Chen, Yun-Nung},
    booktitle   = {Proceedings of the 2024 Conference on Empirical Methods in Natural Language Processing},
    month       = nov,
    year        = {2024},
    address     = {Miami, Florida, USA},
    publisher   = {Association for Computational Linguistics},
    url         = {https://aclanthology.org/2024.emnlp-main.896/},
    doi         = {10.18653/v1/2024.emnlp-main.896},
    pages       = {16028--16045},
}

@inproceedings{Jourdan:2025,
    title       = {Identifying Reliable Evaluation Metrics for Scientific Text Revision},
    author      = {Jourdan, Leane and Hernandez, Nicolas and Boudin, Florian and Dufour, Richard},
    editor      = {Che, Wanxiang and Nabende, Joyce and Shutova, Ekaterina and Pilehvar, Mohammad Taher},
    booktitle   = {Proceedings of the 63rd Annual Meeting of the Association for Computational Linguistics (Volume 1: Long Papers)},
    month       = jul,
    year        = {2025},
    address     = {Vienna, Austria},
    publisher   = {Association for Computational Linguistics},
    url         = {https://aclanthology.org/2025.acl-long.335/},
    pages       = {6731--6756},
}

@inproceedings{Liu:2023,
    title       = {{G}-Eval: {NLG} Evaluation using Gpt-4 with Better Human Alignment},
    author      = {Liu, Yang and Iter, Dan and Xu, Yichong and Wang, Shuohang and Xu, Ruochen  and Zhu, Chenguang},
    editor      = {Bouamor, Houda and Pino, Juan and Bali, Kalika},
    booktitle   = {Proceedings of the 2023 Conference on Empirical Methods in Natural Language Processing},
    month       = dec,
    year        = {2023},
    address     = {Singapore},
    publisher   = {Association for Computational Linguistics},
    url         = {https://aclanthology.org/2023.emnlp-main.153/},
    doi         = {10.18653/v1/2023.emnlp-main.153},
    pages       = {2511--2522},
}

@inproceedings{Zheng:2023,
    author      = {Zheng, Lianmin and Chiang, Wei-Lin and Sheng, Ying and Zhuang, Siyuan and Wu, Zhanghao and Zhuang, Yonghao and Lin, Zi and Li, Zhuohan and Li, Dacheng and Xing, Eric and Zhang, Hao and Gonzalez, Joseph E and Stoica, Ion},
    booktitle   = {Advances in Neural Information Processing Systems},
    editor      = {A. Oh and T. Naumann and A. Globerson and K. Saenko and M. Hardt and S. Levine},
    pages       = {46595--46623},
    publisher   = {Curran Associates, Inc.},
    title       = {Judging LLM-as-a-Judge with MT-Bench and Chatbot Arena},
    url         = {https://proceedings.neurips.cc/paper_files/paper/2023/file/91f18a1287b398d378ef22505bf41832-Paper-Datasets_and_Benchmarks.pdf},
    volume      = {36},
    year        = {2023}
}

@inproceedings{Afzal:2026,
    title       = {Beyond ``Not Novel Enough'': Enriching Scholarly Critique with {LLM}-Assisted Feedback},
    author      = {Afzal, Osama Mohammed and Nakov, Preslav and Hope, Tom and Gurevych, Iryna},
    booktitle   = {Proceedings of the 19th Conference of the {E}uropean Chapter of the {A}ssociation for {C}omputational {L}inguistics (Volume 1: Long Papers)},
    month       = mar,
    year        = {2026},
    address     = {Rabat, Morocco},
    publisher   = {Association for Computational Linguistics},
    url         = {https://aclanthology.org/2026.eacl-long.121/},
    doi         = {10.18653/v1/2026.eacl-long.121},
    pages       = {2648--2671},
}

@misc{Openai:2025,
  author       = {OpenAI},
  title        = {GPT-5.1: A smarter, more conversational ChatGPT},
  year         = {2025},
  howpublished = {\url{https://openai.com/index/gpt-5-1/}},
  note         = {Accessed: 2025-12-11}
}

@article{Deepseekai:2025,
    title       = {DeepSeek-R1 Incentivizes Reasoning in LLMs through reinforcement learning}, 
    author      = {Guo, Daya and Yang, Dejian and Zhang, Haowei and others},
    year        = {2025},
    journal     = {Nature},
    url         = {https://doi.org/10.1038/s41586-025-09422-z},
    volume      = {645},
    pages       = {633--638},
    doi         = {10.1038/s41586-025-09422-z},
}

@misc{IBM:2025,
  author       = {IBM},
  title        = {Granite-3.3-8B-Instruct},
  year         = {2025},
  howpublished = {\url{https://huggingface.co/ibm-granite/granite-3.3-8b-instruct}},
  note         = {Accessed: 2025-12-11}
}

@misc{Grattafiori:2024,
    title       = {The Llama 3 Herd of Models}, 
    author      = {Aaron Grattafiori and Abhimanyu Dubey and Abhinav Jauhri and Abhinav Pandey and Abhishek Kadian and others},
    year        = {2024},
    eprint      = {2407.21783},
    archivePrefix={arXiv},
    primaryClass={cs.AI},
    url         = {https://arxiv.org/abs/2407.21783}, 
}

@misc{Yang:2025,
    title       = {Qwen3 Technical Report}, 
    author={An Yang and Anfeng Li and Baosong Yang and Beichen Zhang and Binyuan Hui and Bo Zheng and Bowen Yu and Chang Gao and Chengen Huang and Chenxu Lv and Chujie Zheng and Dayiheng Liu and Fan Zhou and Fei Huang and Feng Hu and Hao Ge and Haoran Wei and Huan Lin and Jialong Tang and Jian Yang and Jianhong Tu and Jianwei Zhang and Jianxin Yang and Jiaxi Yang and Jing Zhou and Jingren Zhou and Junyang Lin and Kai Dang and Keqin Bao and Kexin Yang and Le Yu and Lianghao Deng and Mei Li and Mingfeng Xue and Mingze Li and Pei Zhang and Peng Wang and Qin Zhu and Rui Men and Ruize Gao and Shixuan Liu and Shuang Luo and Tianhao Li and Tianyi Tang and Wenbiao Yin and Xingzhang Ren and Xinyu Wang and Xinyu Zhang and Xuancheng Ren and Yang Fan and Yang Su and Yichang Zhang and Yinger Zhang and Yu Wan and Yuqiong Liu and Zekun Wang and Zeyu Cui and Zhenru Zhang and Zhipeng Zhou and Zihan Qiu},
    year        = {2025},
    eprint      = {2505.09388},
    archivePrefix={arXiv},
    primaryClass={cs.CL},
    url         = {https://arxiv.org/abs/2505.09388}, 
}

@misc{Liu:2025dsgrm,
    title       = {Inference-Time Scaling for Generalist Reward Modeling}, 
    author      = {Zijun Liu and Peiyi Wang and Runxin Xu and Shirong Ma and Chong Ruan and Peng Li and Yang Liu and Yu Wu},
    year        = {2025},
    eprint      = {2504.02495},
    archivePrefix={arXiv},
    primaryClass={cs.CL},
    url         = {https://arxiv.org/abs/2504.02495},
}

@inproceedings{Kim:2024,
    title       = {Prometheus 2: An Open Source Language Model Specialized in Evaluating Other Language Models},
    author      = {Kim, Seungone and Suk, Juyoung and Longpre, Shayne and Lin, Bill Yuchen and Shin, Jamin and Welleck, Sean and Neubig, Graham and Lee, Moontae and Lee, Kyungjae and Seo, Minjoon},
    booktitle   = {Proceedings of the 2024 Conference on Empirical Methods in Natural Language Processing},
    month       = nov,
    year        = {2024},
    address     = {Miami, Florida, USA},
    publisher   = {Association for Computational Linguistics},
    url         = {https://aclanthology.org/2024.emnlp-main.248/},
    doi         = {10.18653/v1/2024.emnlp-main.248},
    pages       = {4334--4353},
}

@misc{Alexandru:2025,
    title       = {Atla Selene Mini: A General Purpose Evaluation Model}, 
    author      = {Andrei Alexandru and Antonia Calvi and Henry Broomfield and Jackson Golden and Kyle Dai and Mathias Leys and Maurice Burger and Max Bartolo and Roman Engeler and Sashank Pisupati and Toby Drane and Young Sun Park},
    year        = {2025},
    eprint      = {2501.17195},
    archivePrefix={arXiv},
    primaryClass={cs.CL},
    url         = {https://arxiv.org/abs/2501.17195}, 
}

@misc{FlowJudge:2024,
    title       = {Flow Judge: An Open Small Language Model for LLM System Evaluations},
    author      = {{Flow AI}},
    howpublished = {\url{https://flow-ai.com/blog/flow-judge}},
    year        = {2024},
    note        = {Technical report, accessed: 2025-12-27},
    url         = {https://flow-ai.com/blog/flow-judge}
}

@misc{Tang:2025,
    title       = {LongRM: Revealing and Unlocking the Context Boundary of Reward Modeling}, 
    author      = {Zecheng Tang and Baibei Ji and Quantong Qiu and Haitian Wang and Xiaobo Liang and Juntao Li and Min Zhang},
    year        = {2025},
    eprint      = {2510.06915},
    archivePrefix={arXiv},
    primaryClass={cs.CL},
    url         = {https://arxiv.org/abs/2510.06915}, 
}

@inproceedings{Guo:2025,
    title       = {Reward Reasoning Models},
    author      = {Jiaxin Guo and Zewen Chi and Li Dong and Qingxiu Dong and Xun Wu and Shaohan Huang and Furu Wei},
    booktitle   = {Thirty-ninth Annual Conference on Neural Information Processing Systems},
    year        = {2025},
    url         = {https://openreview.net/forum?id=V8Kbz7l2cr}
}

@misc{Ankner:2024,
    title       = {Critique-out-Loud Reward Models}, 
    author      = {Zachary Ankner and Mansheej Paul and Brandon Cui and Jonathan D. Chang and Prithviraj Ammanabrolu},
    year        = {2024},
    eprint      = {2408.11791},
    archivePrefix={arXiv},
    primaryClass={cs.LG},
    url         = {https://arxiv.org/abs/2408.11791}, 
}

@misc{Jian:2025,
    title       = {PaTaRM: Bridging Pairwise and Pointwise Signals via Preference-Aware Task-Adaptive Reward Modeling}, 
    author      = {Ai Jian and Jingqing Ruan and Xing Ma and Dailin Li and QianLin Zhou and Ke Zeng and Xunliang Cai},
    year        = {2025},
    eprint      = {2510.24235},
    archivePrefix={arXiv},
    primaryClass={cs.LG},
    url         = {https://arxiv.org/abs/2510.24235}, 
}

@inproceedings{Chen:2026,
    title       = {{RM}-R1: Reward Modeling as Reasoning},
    author      = {Xiusi Chen and Gaotang Li and Ziqi Wang and Bowen Jin and Cheng Qian and Yu Wang and Hongru WANG and Yu Zhang and Denghui Zhang and Tong Zhang and Hanghang Tong and Heng Ji},
    booktitle   = {Fourteenth International Conference on Learning Representations},
    year        = {2026},
    url         = {https://openreview.net/forum?id=1ZqJ6jj75q}
}

@misc{Chen:2025judgelrm,
    title       = {JudgeLRM: Large Reasoning Models as a Judge}, 
    author      = {Nuo Chen and Zhiyuan Hu and Qingyun Zou and Jiaying Wu and Qian Wang and Bryan Hooi and Bingsheng He},
    year        = {2025},
    eprint      = {2504.00050},
    archivePrefix={arXiv},
    primaryClass={cs.CL},
    url         = {https://arxiv.org/abs/2504.00050}, 
}

@inproceedings{Wen:2026,
    title       = {Reinforcement Learning with Verifiable Rewards Implicitly Incentivizes Correct Reasoning in Base {LLM}s},
    author      = {Xumeng Wen and Zihan Liu and Shun Zheng and Shengyu Ye and Zhirong Wu and Yang Wang and Zhijian Xu and Xiao Liang and Junjie Li and Ziming Miao and Jiang Bian and Mao Yang},
    booktitle   = {Fourteenth International Conference on Learning Representations},
    year        = {2026},
    url         = {https://openreview.net/forum?id=jGbRWwIidy}
}

@misc{Wang:2024,
    title       = {Self-Taught Evaluators}, 
    author      = {Tianlu Wang and Ilia Kulikov and Olga Golovneva and Ping Yu and Weizhe Yuan and Jane Dwivedi-Yu and Richard Yuanzhe Pang and Maryam Fazel-Zarandi and Jason Weston and Xian Li},
    year        = {2024},
    eprint      = {2408.02666},
    archivePrefix={arXiv},
    primaryClass={cs.CL},
    url         = {https://arxiv.org/abs/2408.02666}, 
}

@misc{Skyworkcritic:2024,
    title       = {Skywork Critic Model Series},
    author      = {Shiwen, Tu and Liang, Zhao and Liu, Chris Yuhao and Zeng, Liang and Liu, Yang},
    year        = {2024},
    month       = {September},
    howpublished={\url{https://huggingface.co/Skywork}},
    url         = {https://huggingface.co/Skywork},
}

@misc{Liu:2025,
    title       = {OpenRubrics: Towards Scalable Synthetic Rubric Generation for Reward Modeling and LLM Alignment}, 
    author      = {Tianci Liu and Ran Xu and Tony Yu and Ilgee Hong and Carl Yang and Tuo Zhao and Haoyu Wang},
    year        = {2025},
    eprint      = {2510.07743},
    archivePrefix={arXiv},
    primaryClass={cs.CL},
    url         = {https://arxiv.org/abs/2510.07743}, 
}

@inproceedings{Liang:2025,
    title       = {Generative Reward Modeling via Synthetic Criteria Preference Learning},
    author      = {Liang, Xiaobo and Zhang, Haoke and Li, Juntao and Chen, Kehai  and Zhu, Qiaoming and Zhang, Min},
    booktitle   = {Proceedings of the 63rd Annual Meeting of the Association for Computational Linguistics (Volume 1: Long Papers)},
    month       = jul,
    year        = {2025},
    address     = {Vienna, Austria},
    publisher   = {Association for Computational Linguistics},
    url         = {https://aclanthology.org/2025.acl-long.1297/},
    doi         = {10.18653/v1/2025.acl-long.1297},
    pages       = {26755--26769},
}

@inproceedings{Li:2024rw-survey,
    title       = {Related Work and Citation Text Generation: A Survey},
    author      = {Li, Xiangci and Ouyang, Jessica},
    editor      = {Al-Onaizan, Yaser and Bansal, Mohit and Chen, Yun-Nung},
    booktitle   = {Proceedings of the 2024 Conference on Empirical Methods in Natural Language Processing},
    month       = nov,
    year        = {2024},
    address     = {Miami, Florida, USA},
    publisher   = {Association for Computational Linguistics},
    url         = {https://aclanthology.org/2024.emnlp-main.767/},
    doi         = {10.18653/v1/2024.emnlp-main.767},
    pages       = {13846--13864},
}

@article{Liang:2024,
    author      = {Weixin Liang and Yuhui Zhang and Hancheng Cao and Binglu Wang and Daisy Yi Ding and Xinyu Yang and Kailas Vodrahalli and Siyu He and Daniel Scott Smith and Yian Yin and Daniel A. McFarland and James Zou},
    title       = {Can Large Language Models Provide Useful Feedback on Research Papers? A Large-Scale Empirical Analysis},
    journal     = {NEJM AI},
    volume      = {1},
    number      = {8},
    pages       = {AIoa2400196},
    year        = {2024},
    doi         = {10.1056/AIoa2400196},
    url         = {https://ai.nejm.org/doi/full/10.1056/AIoa2400196},
}

@inproceedings{Yang:2024,
    author      = {Yang, Rui and Ding, Ruomeng and Lin, Yong and Zhang, Huan and Zhang, Tong},
    booktitle   = {Advances in Neural Information Processing Systems},
    doi         = {10.52202/079017-1990},
    pages       = {62279--62309},
    publisher   = {Curran Associates, Inc.},
    title       = {Regularizing Hidden States Enables Learning Generalizable Reward Model for LLMs},
    url         = {https://proceedings.neurips.cc/paper_files/paper/2024/file/71f7154547c748c8041505521ca433ab-Paper-Conference.pdf},
    volume      = {37},
    year        = {2024}
}

@misc{Lai:2024,
    title       = {Step-DPO: Step-wise Preference Optimization for Long-chain Reasoning of LLMs}, 
    author      = {Xin Lai and Zhuotao Tian and Yukang Chen and Senqiao Yang and Xiangru Peng and Jiaya Jia},
    year        = {2024},
    eprint      = {2406.18629},
    archivePrefix={arXiv},
    primaryClass={cs.LG},
    url         = {https://arxiv.org/abs/2406.18629}, 
}

@inproceedings{Chakrabarty:2025,
    title       = {{AI}-Slop to {AI}-Polish? Aligning Language Models through Edit-Based Writing Rewards and Test-time computation},
    author      = {Tuhin Chakrabarty and Philippe Laban and Chien-Sheng Wu},
    booktitle   = {Second Conference on Language Modeling},
    year        = {2025},
    url         = {https://openreview.net/forum?id=jeDYcjuZIV}
}

\appendix

\section{Dataset} \label{app:dataset}

Since RevUtil dataset contains both human-annotated and synthetically generated labels, we merge these two splits. For the \textit{Verifiability} aspect, we divide the evaluation into two sub-evaluations. We first determine whether the given review passage contains any claims under the \textit{Verifiability extraction} evaluation. In the next step, we evaluate the degree of verifiability of data instances that accommodate claims. We provide the number of instances and the corresponding scoring rubrics for related work and RevUtil datasets in Tables \ref{tab:rw-data} and \ref{tab:revutil-data}, respectively. In the final dataset, we combine these two sources, resulting in 65,357 data instances, with 58,712 used for training and 6,645 reserved for testing.

\begin{table}[ht]
\centering
\begin{tabular}{lccc}
\hline
\textbf{Rel. Work} & \textbf{Train} & \textbf{Test} & \textbf{Scoring} \\
\hline
Positioning Type & 954  & 204  & 0--1 \\
Positioning Cons. & 2,822 & 605  & 0--1 \\
Coherence & 4,890 & 1,048 & 0--1 \\
\hline
\textbf{Total} & \textbf{8,666} & \textbf{1,857} & \textbf{10,523} \\
\hline
\end{tabular}
\caption{Statistics and scoring rubric of each aspect in related work dataset.}
\label{tab:rw-data}
\end{table}

\begin{table}[ht]
\centering
\begin{tabular}{lccc}
\hline
\textbf{RevUtil} & \textbf{Train} & \textbf{Test} & \textbf{Scoring} \\
\hline
Actionability & 10,432  & 1,000  & 1--5 \\
Grounding & 10,431 & 1,000  & 1--5 \\
Verifiability Ext. & 10,430 & 1,000 & 0--1 \\
Verifiability & 8,323 & 788 & 1--5 \\
Helpfulness & 10,430 & 1,000 & 1--5 \\
\hline
\textbf{Total} & \textbf{50,046} & \textbf{4,788} & \textbf{54,834} \\
\hline
\end{tabular}
\caption{Statistics and scoring rubric of each aspect in RevUtil dataset.}
\label{tab:revutil-data}
\end{table}

\section{Experimental Details} \label{app:exp}

We use the Unsloth \cite{Unsloth:2023} framework for LoRA optimization. We select LoRA rank as 64. In the roll-out phase, we utilize vLLM framework\footnote{https://docs.vllm.ai/en/stable/} while generating candidate responses. For each data instance, we sample 4 prompts via nucleus sampling with temperature of 1 and top-p of 0.95. Training is performed on a single NVIDIA A100 GPU (80 GB) and takes approximately three days.

\section{Evaluation Prompts} \label{app:prompts}

We provide the list of prompts used while evaluating scientific writings in Table \ref{tab:prompts}.

\begin{table}[h]
\centering
\begin{tabular}{l l}
\hline
\textbf{Evaluation Type} & \textbf{Prompts} \\
\hline
System prompt & Figure \ref{fig:prompt_system} \\
RW - Coherence & Figures \ref{fig:prompt_rw_coherence_part_1}, \ref{fig:prompt_rw_coherence_part_2} \\
RW - Pos. Type & Figures \ref{fig:prompt_rw_pos_type_each_part_1}, \ref{fig:prompt_rw_pos_type_each_part_2}, \ref{fig:prompt_rw_pos_type_each_part_3}, \ref{fig:prompt_rw_pos_type_final} \\
RW - Pos. Cons. & Figures \ref{fig:prompt_rw_pos_cons_each_part_1}, \ref{fig:prompt_rw_pos_cons_each_part_2}, \ref{fig:prompt_rw_pos_cons_final_part_1}, \ref{fig:prompt_rw_pos_cons_final_part_2} \\
RU - Verifiability Ext. & Figure \ref{fig:prompt_ru_ver_ext} \\
RU - Verifiability & Figures \ref{fig:prompt_ru_verifiability_part_1}, \ref{fig:prompt_ru_verifiability_part_2} \\
RU - Actionability & Figures \ref{fig:prompt_ru_actionability_part_1}, \ref{fig:prompt_ru_actionability_part_2} \\
RU - Grounding Spec. & Figures \ref{fig:prompt_ru_grounding_part_1}, \ref{fig:prompt_ru_grounding_part_2} \\
RU - Helpfulness & Figures \ref{fig:prompt_ru_helpfulness_part_1}, \ref{fig:prompt_ru_helpfulness_part_2} \\
Nov. Alignment & Figures \ref{fig:prompt_novelty_part_1}, \ref{fig:prompt_novelty_part_2} \\
Rev. - Relatedness & Figures \ref{fig:prompt_revision_rel_part_1}, \ref{fig:prompt_revision_rel_part_2} \\
Rev. - Correctness & Figures \ref{fig:prompt_revision_cor_part_1}, \ref{fig:prompt_revision_cor_part_2} \\
\hline
\end{tabular}
\caption{Evaluation prompts table}
\label{tab:prompts}
\end{table}

\section{Full Results} \label{app:full_results}

We provide full numeric results in Tables \ref{tab:rw_gen_full}, \ref{tab:revutil_full}, \ref{tab:unseen_full} for accuracy and \ref{tab:rw_gen_full_f1}, \ref{tab:revutil_full_f1}, \ref{tab:unseen_full_f1} for F1 Score.

\begin{table*}[ht]
\centering
\begin{tabular}{l|cccc}
\hline

\textbf{Rel. Work} & \textbf{Coherence} & \textbf{Positioning Cons.} & \textbf{Positioning Type} & \textbf{Overall Task} \\
\hline
DS-Qwen & 0.61 $\pm$ 0.010 & 0.85 $\pm$ 0.010 & 0.66 $\pm$ 0.017 & 0.70 $\pm$ 0.007 \\

Granite3.3 & 0.66 $\pm$ 0.007 & 0.88 $\pm$ 0.006 & 0.63 $\pm$ 0.016 & 0.73 $\pm$ 0.005 \\

Llama3.1 & 0.62 $\pm$ 0.013 & 0.89 $\pm$ 0.010 & 0.60 $\pm$ 0.029 & 0.70 $\pm$ 0.012 \\

Qwen2.5 & 0.67 $\pm$ 0.010 & 0.94 $\pm$ 0.006 & 0.68 $\pm$ 0.018 & 0.76 $\pm$ 0.008 \\

Qwen3 & 0.65 $\pm$ 0.003 & 0.95 $\pm$ 0.009 & 0.96 $\pm$ 0.009 & 0.78 $\pm$ 0.003 \\
\hline

Prometheus & 0.64 $\pm$ 0.012 & 0.87 $\pm$ 0.011 & 0.61 $\pm$ 0.017 & 0.71 $\pm$ 0.007 \\

Selene & 0.63 $\pm$ 0.010 & 0.88 $\pm$ 0.004 & 0.59 $\pm$ 0.016 & 0.71 $\pm$ 0.005 \\

Skywork & 0.69 $\pm$ 0.005 & 0.94 $\pm$ 0.006 & 0.78 $\pm$ 0.030 & 0.78 $\pm$ 0.005 \\

DS-GRM & 0.57 $\pm$ 0.012 & 0.85 $\pm$ 0.012 & 0.71 $\pm$ 0.011 & 0.68 $\pm$ 0.010 \\
\hline

o3-mini & 0.63 $\pm$ 0.003 & 0.99 $\pm$ 0.002 & 0.998 $\pm$ 0.002 & 0.79 $\pm$ 0.002 \\

GPT-5.2 & 0.73 $\pm$ 0.002 & 0.94 $\pm$ 0.004 & 1.00 $\pm$ 0.000 & 0.83 $\pm$ 0.002 \\
\hline

Sci-RM & 0.73 $\pm$ 0.007 & 0.995 $\pm$ 0.001 & 0.86 $\pm$ 0.011 & 0.83 $\pm$ 0.005 \\

Sci-RM Ref. & 0.73 $\pm$ 0.004 & 0.996 $\pm$ 0.001 & 0.87 $\pm$ 0.012 & 0.83 $\pm$ 0.003 \\
\hline
\end{tabular}
\caption{Full accuracy results for related work evaluation. Values are reported as mean $\pm$ standard deviation.}
\label{tab:rw_gen_full}
\end{table*}

\begin{table*}[ht]
\centering
\small
\begin{tabular}{l|cccccc}
\hline
\textbf{RevUtil} & \textbf{Verifiability Ext.} & \textbf{Verifiability} & \textbf{Actionability} &  \textbf{Grounding Spec.} &  \textbf{Helpfulness} & \textbf{Overall Task} \\
\hline
DS-Qwen & 0.81 $\pm$ 0.005 & 0.39 $\pm$ 0.017 & 0.30 $\pm$ 0.006 & 0.34 $\pm$ 0.004 & 0.36 $\pm$ 0.014 & 0.44 $\pm$ 0.003 \\

Granite3.3 & 0.82 $\pm$ 0.006 & 0.39 $\pm$ 0.011 & 0.32 $\pm$ 0.004 & 0.40 $\pm$ 0.011 & 0.36 $\pm$ 0.009 & 0.46 $\pm$ 0.005 \\

Llama3.1 & 0.87 $\pm$ 0.004 & 0.27 $\pm$ 0.006 & 0.30 $\pm$ 0.009 & 0.43 $\pm$ 0.015 & 0.42 $\pm$ 0.007 & 0.47 $\pm$ 0.004 \\

Qwen2.5 & 0.85 $\pm$ 0.007 & 0.38 $\pm$ 0.016 & 0.32 $\pm$ 0.008 & 0.40 $\pm$ 0.006 & 0.45 $\pm$ 0.014 & 0.49 $\pm$ 0.005 \\

Qwen3 & 0.88 $\pm$ 0.005 & 0.50 $\pm$ 0.012 & 0.42 $\pm$ 0.010 & 0.60 $\pm$ 0.005 & 0.43 $\pm$ 0.010 & 0.57 $\pm$ 0.005 \\
\hline

Prometheus & 0.80 $\pm$ 0.015 & 0.40 $\pm$ 0.006 & 0.36 $\pm$ 0.017 & 0.29 $\pm$ 0.013 & 0.40 $\pm$ 0.008 & 0.45 $\pm$ 0.010 \\

Selene & 0.87 $\pm$ 0.003 & 0.44 $\pm$ 0.005 & 0.32 $\pm$ 0.015 & 0.40 $\pm$ 0.005 & 0.46 $\pm$ 0.010 & 0.50 $\pm$ 0.005 \\

Skywork & 0.90 $\pm$ 0.006 & 0.42 $\pm$ 0.019 & 0.43 $\pm$ 0.010 & 0.48 $\pm$ 0.011 & 0.42 $\pm$ 0.010 & 0.53 $\pm$ 0.005 \\

DS-GRM & 0.70 $\pm$ 0.009 & 0.39 $\pm$ 0.011 & 0.35 $\pm$ 0.015 & 0.39 $\pm$ 0.010 & 0.41 $\pm$ 0.008 & 0.45 $\pm$ 0.005 \\
\hline

o3-mini & 0.86 $\pm$ 0.002 & 0.48 $\pm$ 0.007 & 0.49 $\pm$ 0.010 & 0.63 $\pm$ 0.007 & 0.51 $\pm$ 0.005 & 0.60 $\pm$ 0.001 \\

GPT-5.2 & 0.89 $\pm$ 0.002 & 0.46 $\pm$ 0.004 & 0.40 $\pm$ 0.008 & 0.66 $\pm$ 0.004 & 0.49 $\pm$ 0.009 & 0.59 $\pm$ 0.004 \\
\hline

Sci-RM & 0.92 $\pm$ 0.002 & 0.65 $\pm$ 0.007 & 0.62 $\pm$ 0.009 & 0.70 $\pm$ 0.006 & 0.63 $\pm$ 0.012 & 0.71 $\pm$ 0.006 \\

Sci-RM Ref. & 0.92 $\pm$ 0.005 & 0.65 $\pm$ 0.007 & 0.58 $\pm$ 0.007 & 0.70 $\pm$ 0.008 & 0.62 $\pm$ 0.003 & 0.69 $\pm$ 0.003 \\
\hline
\end{tabular}
\caption{Full accuracy results for review utility evaluation. Values are reported as mean $\pm$ standard deviation.}
\label{tab:revutil_full}
\end{table*}

\begin{table*}[ht]
\centering
\begin{tabular}{l|c|ccc}
\hline
\multirow{2}{*}{\textbf{Model}} & \textbf{Novelty} & \multicolumn{3}{c}{\textbf{Revision}} \\
& \textbf{Alignment} & \textbf{Relatedness} & \textbf{Correctness} & \textbf{Overall Task} \\
\hline
DS-Qwen & 0.41 $\pm$ 0.07 & 0.77 $\pm$ 0.001 & 0.67 $\pm$ 0.005 & 0.72 $\pm$ 0.002 \\

Granite3.3 & 0.48 $\pm$ 0.04 & 0.80 $\pm$ 0.003 & 0.68 $\pm$ 0.003 & 0.74 $\pm$ 0.003 \\

Llama3.1 & 0.56 $\pm$ 0.04 & 0.56 $\pm$ 0.009 & 0.61 $\pm$ 0.006 & 0.58 $\pm$ 0.005 \\

Qwen2.5 & 0.71 $\pm$ 0.05 & 0.80 $\pm$ 0.005 & 0.66 $\pm$ 0.010 & 0.73 $\pm$ 0.006 \\

Qwen3 & 0.58 $\pm$ 0.05 & 0.86 $\pm$ 0.004 & 0.76 $\pm$ 0.003 & 0.81 $\pm$ 0.003 \\
\hline

Prometheus & 0.33 $\pm$ 0.06 & 0.76 $\pm$ 0.006 & 0.70 $\pm$ 0.003 & 0.73 $\pm$ 0.002 \\

Selene & 0.57 $\pm$ 0.04 & 0.79 $\pm$ 0.004 & 0.71 $\pm$ 0.004 & 0.75 $\pm$ 0.003 \\

Skywork & 0.42 $\pm$ 0.05 & 0.83 $\pm$ 0.003 & 0.77 $\pm$ 0.004 & 0.80 $\pm$ 0.002 \\

DS-GRM & 0.35 $\pm$ 0.05 & 0.73 $\pm$ 0.001 & 0.59 $\pm$ 0.008 & 0.66 $\pm$ 0.004 \\
\hline

o3-mini & 0.78 $\pm$ 0.01 & 0.88 $\pm$ 0.002 & 0.80 $\pm$ 0.002 & 0.84 $\pm$ 0.0008 \\

GPT-5.2 & 0.74 $\pm$ 0.06 & 0.82 $\pm$ 0.003 & 0.78 $\pm$ 0.002 & 0.80 $\pm$ 0.001 \\
\hline

Sci-RM & 0.61 $\pm$ 0.04 & 0.82 $\pm$ 0.001 & 0.73 $\pm$ 0.005 & 0.78 $\pm$ 0.002 \\

Sci-RM Ref. & 0.74 $\pm$ 0.04 & 0.82 $\pm$ 0.002 & 0.73 $\pm$ 0.003 & 0.78 $\pm$ 0.002 \\
\hline
\end{tabular}
\caption{Full accuracy results for unseen novelty and revision evaluation tasks. Values are reported as mean $\pm$ standard deviation.}
\label{tab:unseen_full}
\end{table*}

\begin{table*}[ht]
\centering
\begin{tabular}{l|cccc}
\hline

\textbf{Rel. Work} & \textbf{Coherence} & \textbf{Positioning Cons.} & \textbf{Positioning Type} & \textbf{Overall Task} \\
\hline
DS-Qwen & 0.61 $\pm$ 0.012 & 0.85 $\pm$ 0.011 & 0.66 $\pm$ 0.019 & 0.71 $\pm$ 0.007 \\

Granite3.3 & 0.66 $\pm$ 0.009 & 0.88 $\pm$ 0.007 & 0.64 $\pm$ 0.019 & 0.73 $\pm$ 0.006 \\

Llama3.1 & 0.63 $\pm$ 0.016 & 0.89 $\pm$ 0.011 & 0.61 $\pm$ 0.031 & 0.71 $\pm$ 0.016 \\

Qwen2.5 & 0.67 $\pm$ 0.012 & 0.95 $\pm$ 0.005 & 0.70 $\pm$ 0.022 & 0.77 $\pm$ 0.009 \\

Qwen3 & 0.65 $\pm$ 0.001 & 0.95 $\pm$ 0.011 & 0.96 $\pm$ 0.008 & 0.86 $\pm$ 0.003 \\
\hline

Prometheus & 0.64 $\pm$ 0.014 & 0.88 $\pm$ 0.013 & 0.63 $\pm$ 0.017 & 0.72 $\pm$ 0.008 \\

Selene & 0.63 $\pm$ 0.011 & 0.88 $\pm$ 0.004 & 0.58 $\pm$ 0.017 & 0.70 $\pm$ 0.003 \\

Skywork & 0.69 $\pm$ 0.006 & 0.94 $\pm$ 0.007 & 0.77 $\pm$ 0.036 & 0.80 $\pm$ 0.013 \\

DS-GRM & 0.58 $\pm$ 0.014 & 0.86 $\pm$ 0.011 & 0.68 $\pm$ 0.019 & 0.71 $\pm$ 0.010 \\
\hline

o3-mini & 0.59 $\pm$ 0.003 & 0.99 $\pm$ 0.002 & 0.998 $\pm$ 0.003 & 0.86 $\pm$ 0.001 \\

GPT-5.2 & 0.72 $\pm$ 0.003 & 0.94 $\pm$ 0.005 & 1.00 $\pm$ 0.000 & 0.89 $\pm$ 0.002 \\
\hline

Sci-RM & 0.73 $\pm$ 0.008 & 0.99 $\pm$ 0.001 & 0.86 $\pm$ 0.013 & 0.86 $\pm$ 0.005 \\

Sci-RM Ref. & 0.73 $\pm$ 0.004 & 0.996 $\pm$ 0.001 & 0.87 $\pm$ 0.013 & 0.87 $\pm$ 0.005 \\
\hline
\end{tabular}
\caption{Full F1 score results for related work evaluation. Values are reported as mean $\pm$ standard deviation.}
\label{tab:rw_gen_full_f1}
\end{table*}

\begin{table*}[ht]
\centering
\small
\begin{tabular}{l|cccccc}
\hline
\textbf{RevUtil} & \textbf{Verifiability Ext.} & \textbf{Verifiability} & \textbf{Actionability} &  \textbf{Grounding Spec.} &  \textbf{Helpfulness} & \textbf{Overall Task} \\
\hline
DS-Qwen & 0.82 $\pm$ 0.004 & 0.42 $\pm$ 0.018 & 0.29 $\pm$ 0.006 & 0.35 $\pm$ 0.004 & 0.38 $\pm$ 0.015 & 0.45 $\pm$ 0.004 \\

Granite3.3 & 0.80 $\pm$ 0.007 & 0.41 $\pm$ 0.014 & 0.34 $\pm$ 0.003 & 0.42 $\pm$ 0.014 & 0.37 $\pm$ 0.010 & 0.47 $\pm$ 0.007 \\

Llama3.1 & 0.87 $\pm$ 0.005 & 0.31 $\pm$ 0.010 & 0.31 $\pm$ 0.015 & 0.41 $\pm$ 0.021 & 0.43 $\pm$ 0.009 & 0.47 $\pm$ 0.005 \\

Qwen2.5 & 0.85 $\pm$ 0.008 & 0.41 $\pm$ 0.017 & 0.32 $\pm$ 0.011 & 0.39 $\pm$ 0.009 & 0.42 $\pm$ 0.015 & 0.48 $\pm$ 0.006 \\

Qwen3 & 0.88 $\pm$ 0.005 & 0.52 $\pm$ 0.012 & 0.44 $\pm$ 0.012 & 0.60 $\pm$ 0.006 & 0.44 $\pm$ 0.011 & 0.57 $\pm$ 0.005 \\
\hline

Prometheus & 0.81 $\pm$ 0.016 & 0.42 $\pm$ 0.008 & 0.36 $\pm$ 0.018 & 0.34 $\pm$ 0.012 & 0.41 $\pm$ 0.011 & 0.47 $\pm$ 0.009 \\

Selene & 0.85 $\pm$ 0.005 & 0.45 $\pm$ 0.003 & 0.26 $\pm$ 0.014 & 0.42 $\pm$ 0.006 & 0.43 $\pm$ 0.010 & 0.48 $\pm$ 0.004 \\

Skywork & 0.89 $\pm$ 0.008 & 0.44 $\pm$ 0.024 & 0.43 $\pm$ 0.011 & 0.48 $\pm$ 0.015 & 0.44 $\pm$ 0.013 & 0.53 $\pm$ 0.007 \\

DS-GRM & 0.73 $\pm$ 0.008 & 0.40 $\pm$ 0.012 & 0.33 $\pm$ 0.019 & 0.41 $\pm$ 0.010 & 0.41 $\pm$ 0.007 & 0.46 $\pm$ 0.006 \\
\hline

o3-mini & 0.83 $\pm$ 0.005 & 0.51 $\pm$ 0.010 & 0.48 $\pm$ 0.016 & 0.61 $\pm$ 0.008 & 0.52 $\pm$ 0.008 & 0.59 $\pm$ 0.002 \\

GPT-5.2 & 0.88 $\pm$ 0.004 & 0.47 $\pm$ 0.009 & 0.37 $\pm$ 0.010 & 0.65 $\pm$ 0.005 & 0.49 $\pm$ 0.011 & 0.57 $\pm$ 0.006 \\
\hline

Sci-RM & 0.92 $\pm$ 0.003 & 0.59 $\pm$ 0.010 & 0.59 $\pm$ 0.010 & 0.67 $\pm$ 0.007 & 0.62 $\pm$ 0.014 & 0.68 $\pm$ 0.007 \\

Sci-RM Ref. & 0.92 $\pm$ 0.006 & 0.59 $\pm$ 0.010 & 0.53 $\pm$ 0.008 & 0.66 $\pm$ 0.008 & 0.60 $\pm$ 0.004 & 0.66 $\pm$ 0.004 \\
\hline
\end{tabular}
\caption{Full F1 score results for review utility evaluation. Values are reported as mean $\pm$ standard deviation.}
\label{tab:revutil_full_f1}
\end{table*}

\begin{table*}[ht]
\centering
\begin{tabular}{l|c|ccc}
\hline
\multirow{2}{*}{\textbf{Model}} & \textbf{Novelty} & \multicolumn{3}{c}{\textbf{Revision}} \\
& \textbf{Alignment} & \textbf{Relatedness} & \textbf{Correctness} & \textbf{Overall Task} \\
\hline
DS-Qwen & 0.50 $\pm$ 0.092 & 0.74 $\pm$ 0.002 & 0.61 $\pm$ 0.008 & 0.67 $\pm$ 0.004 \\

Granite3.3 & 0.51 $\pm$ 0.041 & 0.75 $\pm$ 0.006 & 0.61 $\pm$ 0.005 & 0.68 $\pm$ 0.005 \\

Llama3.1 & 0.61 $\pm$ 0.043 & 0.61 $\pm$ 0.009 & 0.61 $\pm$ 0.007 & 0.61 $\pm$ 0.005 \\

Qwen2.5 & 0.69 $\pm$ 0.063 & 0.79 $\pm$ 0.006 & 0.65 $\pm$ 0.007 & 0.72 $\pm$ 0.005 \\

Qwen3 & 0.61 $\pm$ 0.054 & 0.86 $\pm$ 0.006 & 0.74 $\pm$ 0.004 & 0.80 $\pm$ 0.004 \\
\hline

Prometheus & 0.39 $\pm$ 0.071 & 0.78 $\pm$ 0.004 & 0.69 $\pm$ 0.004 & 0.74 $\pm$ 0.002 \\

Selene & 0.62 $\pm$ 0.046 & 0.79 $\pm$ 0.004 & 0.70 $\pm$ 0.005 & 0.75 $\pm$ 0.003 \\

Skywork & 0.43 $\pm$ 0.060 & 0.83 $\pm$ 0.003 & 0.77 $\pm$ 0.005 & 0.80 $\pm$ 0.002 \\

DS-GRM & 0.30 $\pm$ 0.091 & 0.76 $\pm$ 0.001 & 0.61 $\pm$ 0.009 & 0.69 $\pm$ 0.004 \\
\hline

o3-mini & 0.78 $\pm$ 0.010 & 0.87 $\pm$ 0.002 & 0.79 $\pm$ 0.002 & 0.83 $\pm$ 0.001 \\

GPT-5.2 & 0.74 $\pm$ 0.010 & 0.83 $\pm$ 0.004 & 0.78 $\pm$ 0.001 & 0.80 $\pm$ 0.001 \\
\hline

Sci-RM & 0.63 $\pm$ 0.046 & 0.81 $\pm$ 0.002 & 0.69 $\pm$ 0.005 & 0.75 $\pm$ 0.003 \\

Sci-RM Ref. & 0.70 $\pm$ 0.040 & 0.81 $\pm$ 0.002 & 0.69 $\pm$ 0.004 & 0.75 $\pm$ 0.003 \\
\hline
\end{tabular}
\caption{Full F1 score results for unseen novelty and revision evaluation tasks. Values are reported as mean $\pm$ standard deviation.}
\label{tab:unseen_full_f1}
\end{table*}

\section{Reasoning Trace Analysis} \label{app:reasoning}

We provide full reasoning trace examples for review utility, novelty alignment evaluation and revision evaluation across different models in Tables \ref{tab:scirm-qwen2.5}, \ref{tab:scirmref-scirm}, and \ref{tab:scirmref-dsgrm}, respectively.

\begin{table*}[ht]
\centering
\renewcommand{\arraystretch}{1.2}
\begin{tabular}{p{0.47\textwidth} p{0.47\textwidth}}
\hline
\textbf{Task:} Review Utility & \textbf{Aspect:} Actionability \\
\hline
\multicolumn{2}{p{0.94\textwidth}}{\textbf{Query Review:} Evaluation datasets are all short outputs: classification or question answering. This is a slightly limited setting for the claims.} \\
\textbf{Gold label:} 1 & \\

\multicolumn{2}{p{0.94\textwidth}}{\textbf{Label 1 Explanation:} The comment lacks meaningful information to help authors improve the paper. Authors do not know what they should do after reading the comment.} \\
\multicolumn{2}{p{0.94\textwidth}}{\textbf{Label 3 Explanation:} The comment explicitly states an action but is vague on how to execute it.} \\

\hline
\textbf{SciRM} & \textbf{Qwen2.5-7B-Instruct} \\
\small <reasoning>The provided comment, "Evaluation datasets are all short outputs: classification or question answering. This is a slightly limited setting for the claims," lacks specific and actionable guidance. It points out a limitation but does not offer a concrete solution or modification. The authors are left to determine how to expand the evaluation datasets, but without more detail, this action is too vague to be considered as a clear and actionable instruction. As a result, this comment is not sufficient for an actionable evaluation score. Therefore, the evaluation score should be 1, indicating that the comment lacks meaningful information to help the authors improve the paper. The reviewer is likely highlighting a limitation and expecting authors to address it, but without providing a clear direction, the comment is not actionable. The provided example would be improved by specifying particular actions or concrete steps that could be taken to improve the dataset or strengthen the claims in the paper.</reasoning>
<score>1</score>
&
\small <reasoning>The comment suggests that the evaluation datasets used in the paper are predominantly short outputs for classification or question answering, implying a limited setting for the claims. It identifies a limitation but doesn't provide specific directions or a clear path for addressing this issue, making it somewhat vague. The authors might infer that they need to use a wider range of datasets, but the comment does not specify how they should go about doing so. Hence, while it is concrete in identifying the limitation, it lacks explicit instructions on how to solve it, resulting in a moderate level of actionability. Therefore, the appropriate score is 3.</reasoning>
<score>3</score> \\
\hline
\end{tabular}
\caption{Reasoning trace comparison for SciRM and base model Qwen2.5-7B-Instruct}
\label{tab:scirm-qwen2.5}
\end{table*}

\begin{table*}[ht]
\centering
\renewcommand{\arraystretch}{1.2}
\begin{tabular}{p{0.47\textwidth} p{0.47\textwidth}}
\hline
\textbf{Task:} Novelty Alignment & \textbf{Aspect:} Coherence \\
\hline
\multicolumn{2}{p{0.94\textwidth}}{\textbf{Query Assessments:} Assessment 1: (...) While the approach leverages transfer learning between 1D and 3D representations and incorporates enhancements in model scaling and architecture, the novelty from a machine learning perspective is limited. The core methodology primarily involves integrating existing components—specifically, LLaMA-based language models and established diffusion models—without introducing fundamentally new techniques. (...) Overall, the contribution is largely an application and combination of known methods rather than a significant advancement in machine learning methodology (...)

Assessment 2: NExT-Mol introduces a novel approach by explicitly transferring knowledge from a large-scale 1D molecular language model (SELFIES-based MoLlama) to a 3D diffusion model for conformer generation, using a cross-modal projector—a combination not previously demonstrated in the literature. While prior works have explored 1D language modeling and 3D diffusion separately or in joint 2D/3D settings, NExT-Mol is the first to leverage large-scale 1D pretraining for 3D molecular generation, resulting in incremental but real improvements on standard 3D benchmarks (FCD, COV-R). The claimed architectural innovations may represent refinements rather than conceptual leaps, and some terminology (e.g., "cross-modal projector") repackages established techniques. The authors' claims of 100\% validity and transformative novelty are somewhat overstated, as 100\% validity is a property of SELFIES and prior models have also scaled up or improved validity. Overall, the main contribution is a substantive integration of established paradigms (large-scale 1D LMs and 3D diffusion) via transfer learning, with advances that are meaningful but largely incremental in the context of a rapidly evolving field.} \\

\textbf{Gold label:} 0 & \\

\multicolumn{2}{p{0.94\textwidth}}{\textbf{Label 0 Explanation:} Two assessments do not come to the same conclusion in terms of novelty verdict.} \\
\multicolumn{2}{p{0.94\textwidth}}{\textbf{Label 1 Explanation:} Two assessments come to the same conclusion in terms of novelty verdict.} \\

\hline
\textbf{SciRM-Ref} & \textbf{SciRM} \\
\small <reasoning>Both assessments emphasize the integration of existing techniques but present different conclusions. Assessment 1 concludes the contribution is not novel, focusing on the lack of fundamental new methodologies and minimal innovation. In contrast, Assessment 2 calls the contribution novel, highlighting the combination of large-scale 1D pretraining for 3D molecular generation as an incremental but real improvement, though still with some limitations in novelty as argued. For the final evaluation, the key is whether both assessments come to the same conclusion regarding the novelty, and in this case, they do not. Assessment 1 considers the contribution non-novel, while Assessment 2 leans towards it being novel but with limited novelty. Therefore, the score should be 0, indicating that the assessments do not align in their final conclusions on the novelty of the paper.
Both assessments provide detailed reasoning for their conclusions, with one suggesting the work lacks fundamental novelty and the other claiming it is an incremental but still novel contribution. The scoring rubric requires a binary judgment on whether the assessments come to the same conclusion about the novelty. Here, given their differing judgments, the score should be 0.</reasoning>
<score>0</score>
&
\small <reasoning>In the first assessment, the paper is considered not novel as the integration of existing models is deemed minimal and not providing significant advancement. The second assessment also indicates novelty, but recognizes a real, incremental improvement from prior work. Both assessments explicitly argue for a certain level of novelty, but they converge in agreeing that the contributions are not fundamental or groundbreaking, rather they represent a meaningful advancement but are incremental. Hence, the assessments come to the same conclusion: the work is not fundamentally novel, but it does offer incremental but significant contributions. Therefore, a score of 1 is appropriate.</reasoning>
<score>1</score> \\
\hline
\end{tabular}
\caption{Reasoning trace comparison for SciRM-Ref and SciRM}
\label{tab:scirmref-scirm}
\end{table*}

\begin{table*}[ht]
\centering
\renewcommand{\arraystretch}{1.2}
\begin{tabular}{p{0.47\textwidth} p{0.47\textwidth}}
\hline
\textbf{Task:} Revision Evaluation & \textbf{Aspect:} Correctness \\
\hline
\multicolumn{2}{p{0.94\textwidth}}{\textbf{Query Revision:} [ORIGINAL TEXT]: Before we deﬁne the diffusion distance, we brieﬂy introduce the intuition behind it: two nodes are considered similar when they diffuse in a similar way through the graph, and therefore when they inﬂuence the other nodes in a similar manner (Fouss et al., 2012). In other words, two nodes are close if they are in the same cluster which has a consistent local structure. More precisely, the diffusion distance at time K between nodes i and j is deﬁned as follows:

[INSTRUCTION]: Remove the ideas which are not particularly essential for the overall paragraph.

[ANSWER]: Before we define the diffusion distance, we briefly introduce the intuition behind it: two nodes are considered similar when they diffuse in a similar way through the graph, and therefore when they influence the other nodes in a similar manner (Fouss et al., 2012). More precisely, the diffusion distance at time K between nodes i and j is defined as follows:
} \\
\textbf{Gold label:} 1 & \\

\multicolumn{2}{p{0.94\textwidth}}{\textbf{Label 0 Explanation:} The revision is not a better version that can replace the original paragraph in the scientific article.} \\
\multicolumn{2}{p{0.94\textwidth}}{\textbf{Label 1 Explanation:} The revision is a better version that should replace the original paragraph in the scientific article.} \\

\hline
\textbf{SciRM-Ref} & \textbf{DS-GRM} \\
\small <reasoning>The instruction asks to remove ideas that are not essential for the overall paragraph. The answer successfully removes the phrase "In other words, two nodes are close if they are in the same cluster which has a consistent local structure," which is not as essential to the definition of diffusion distance as the other parts of the text. The resulting text maintains the essential information for the definition of diffusion distance and is a better version of the paragraph. Therefore, the evaluation score should be 1, indicating that the revision is a better version that could replace the original paragraph.
In this case, the revised text focuses on the key points of the paragraph without any non-essential ideas, making it a stronger version that meets the criteria. This removal leads to a clearer, more concise definition without losing any critical information, ensuring that the revised version is a good replacement for the original.</reasoning>
<score>1</score>
&
\small <reasoning>The instruction asks to remove the ideas which are not particularly essential for the overall paragraph. The provided answer keeps the main idea of the paragraph, which is introducing the diffusion distance and its relation to the similarity of nodes. However, it does not remove any non-essential details. The revised text is still informative and relevant, but it could be more concise by removing the phrase "in other words" and the subsequent explanation about the local structure of clusters, which are not particularly essential for the overall paragraph. Therefore, the evaluation score should be 0.</reasoning>
<score>0</score>
 \\
\hline
\end{tabular}
\caption{Reasoning trace comparison for SciRM-Ref and DS-GRM baseline}
\label{tab:scirmref-dsgrm}
\end{table*}

\begin{figure*}[h]
\centering
\begin{promptbox}[title=System Prompt, breakable=false]
You are an evaluator of expert-domain scientific writing. You will get a query-answer pair along with criteria explaining the specific evaluation aspect and the scoring rubric. You should evaluate whether the answer satisfy the query based on the given criteria. In addition, examples demonstrating how the evaluation should be performed will be provided. First output your reasoning enclosed between <reasoning> and </reasoning>. Then output your score enclosed between <score> and </score>. Inside <score> provide only the numeric score and nothing else.
\end{promptbox}
\caption{System prompt that is commonly used across each evaluation task and aspect.}
\label{fig:prompt_system}
\end{figure*}

\begin{figure*}
\centering
\begin{promptbox}[title=Prompt for Related Work Evaluation - Coherence (Part 1), breakable=false]
\textbf{[QUERY]:} Given a context from a scientific paper, your task is to write a coherent citation sentence that cites the given paper with a specific citation number. Citation number will be also provided along with the context.\medskip

\textbf{[CRITERIA]:} The paper context supports (entails) the citation sentence. In cases where more than one paper is referenced in the sentence, as long as context in which given citation number fits the paper content, it should be count as entailment as well. In multiple citation cases, the paper does not have to entail whole sentence. Scoring rubric is as follows:\medskip

0: The given context does not align with the context in which the citation number appears in the sentence, or with the overall meaning of the sentence itself.\medskip

1: The given context is compatible with the context in which the citation number appears in the sentence.\medskip

\textbf{[EXAMPLES]:} \medskip

<START OF EXAMPLE 1> \medskip

\textbf{CONTEXT:} \{\textit{Abstract and Introduction of example cited paper}\}

\textbf{CITATION NUMBER:} 10

\textbf{ANSWER:} Additionally, simple predictors that regress on features extracted from architectures, such as textual encoding schemes or direct structural descriptions, have been found effective for performance approximation, enabling significant sample efficiency improvements [9][10]. 

\textbf{EVALUATION:} <reasoning>The context discusses the effectiveness of a Neural Predictor method that uses regression modeling to predict validation accuracy based on architecture features, achieving significant sample efficiency improvements over traditional approaches. The citation sentence aligns with the context in pointing out the effectiveness of simple predictors for performance approximation, which correlates with the regression steps described in the Neural Predictor method. Therefore, the citation sentence is entailed by the context and follows from it. Therefore, the evaluation score should be 1.</reasoning><score>1</score> \medskip

<END OF EXAMPLE 1> \medskip

\end{promptbox}
\caption{Prompt for Related Work Evaluation - Coherence (Part 1)}
\label{fig:prompt_rw_coherence_part_1}
\end{figure*}

\begin{figure*}
\centering
\begin{promptbox}[title=Prompt for Related Work Evaluation - Coherence (Part 2), breakable=false]
<START OF EXAMPLE 2> \medskip

\textbf{CONTEXT:} \{\textit{Abstract and Introduction of example cited paper}\}

\textbf{CITATION NUMBER:} 11

\textbf{ANSWER:} Natural Language Inference has emerged as an essential tool for modeling semantic relationships, with applications extending from textual entailment assessments [11,12] to dialogue consistency verification [17]. 

\textbf{EVALUATION:} <reasoning>The citation sentence discusses Natural Language Inference (NLI) and its applications, which is not directly related or entailed by the presented context on dialogue generation models and their improvements. The context does not specifically mention or address NLI or its use in dialogue consistency verification. As such, the hypothesis cannot be inferred or entailed from the given context. Therefore, the evaluation score should be 0.</reasoning><score>0</score> \medskip

<END OF EXAMPLE 2> \medskip

\textbf{[CONTEXT]:} \{\textit{Abstract and Introduction of corresponding cited paper}\} \medskip

\textbf{[ANSWER]:} \{\textit{Data instance to be evaluated}\}

\end{promptbox}
\caption{Prompt for Related Work Evaluation - Coherence (Part 2)}
\label{fig:prompt_rw_coherence_part_2}
\end{figure*}

\begin{figure*}
\centering
\begin{promptbox}[title=Prompt for Related Work Evaluation - Positioning Type (Each Paragraph) (Part 1), breakable=false]
\textbf{[QUERY]:} Your task is to write related work section for a scientific paper that states the paper's contribution or its position among the literature in each paragraph.\medskip

\textbf{[CRITERIA]:} In a multiple paragraph related work section, every single paragraph should state the contribution of the paper and/or its position among the literature. Ensure that contribution and/or positioning statements should be aligned with the specific focus of each paragraph. Scoring rubric is as follows:\medskip

0: The answer fails to state contribution of the paper and/or its position among the literature in each related work paragraph.\medskip

1: The answer states contribution of the paper and/or its position among the literature in each related work paragraph.\medskip

\textbf{[EXAMPLES]:} \medskip

<START OF EXAMPLE 1> \medskip

\textbf{ANSWER:} Previous studies on TDMR (Task-Dataset-Metric-Result) extraction primarily rely on data sources such as NLP-progress or PapersWithCode. While these sources are popular, they lack rigorous quality control, including standardized TDM entity representation and complete publication coverage. Some efforts, such as [1] and [2], go further by extracting TDM triples and result values, and normalizing them for leaderboard construction. However, these approaches are constrained by their reliance on a closed domain, requiring extracted triples to match a pre-defined TDM taxonomy. Other studies, such as [3] and [4], only extract TDM triples without result values, thereby forgoing leaderboard construction. Meanwhile, [6] focuses on extracting result values, but only in the context of pre-defined TDM triples, making it similar in limitation to [1]. Consequently, these methods are not adaptable to the dynamic nature of emerging benchmarks, where new tasks, datasets, and metrics are frequently introduced.\medskip

In a parallel research thread, scientific leaderboards have been treated as a form of scientific knowledge graphs composed of Task, Dataset, and Metric entities. Information extraction in this context has been widely studied, as seen in works such as [5, 7, 8, 9, 10]. These studies typically focus on linking scientific content to structured entities but often assume that the schema or triples are known in advance.\medskip

In contrast to prior work, our approach addresses several critical limitations. We construct our dataset manually from publications to ensure comprehensive TDMR annotations, rather than relying on incomplete or unstandardized third-party sources. We normalize TDMR tuples for accurate leaderboard construction and propose novel experimental settings that reflect realistic, dynamic scenarios, including cold-start conditions without any pre-defined TDM taxonomy. By simultaneously constructing a scientific knowledge graph and linking publications to it in an open-domain setting, our method is inherently adaptable to the evolving research landscape.

\textbf{EVALUATION:} <reasoning>The related work draft does not mention the main paper's contributions and its position in each individual paragraph; instead, it waits until the final paragraph to outline the specific contributions of the study. The contributions are clearly summarized and detailed only in the final paragraph. Therefore, the evaluation score should be 0.</reasoning><score>0</score> \medskip

<END OF EXAMPLE 1> \medskip

\end{promptbox}
\caption{Prompt for Related Work Evaluation - Positioning Type (Each Paragraph) (Part 1)}
\label{fig:prompt_rw_pos_type_each_part_1}
\end{figure*}

\begin{figure*}
\centering
\begin{promptbox}[title=Prompt for Related Work Evaluation - Positioning Type (Each Paragraph) (Part 2), breakable=false]
<START OF EXAMPLE 2> \medskip

\textbf{ANSWER:} In terms of data source, previous studies use either NLP-progress or paperswithcode. These sources, however, lack rigorous quality assurance, such as standardizing TDM entities across different leaderboards and ensuring complete coverage of relevant publications. Similar to our work, [1] and [2] extract TDM triples along with the results values and apply normalization for leaderboard construction. However, both studies assume a closed domain and match extracted TDM triples to a pre-defined TDM triple taxonomy. On the other hand, some studies only partially extract TDMR tuples and do not apply normalization. For example, [3] and [4] extract TDM triples without results. Therefore, these works do not deal with leaderboard construction. In addition, [6] extract the results values depending on the pre-defined TDM triples. Both [3] and [6] leverage pre-defined TDM triples in an extraction process similar to [1]. Since these approaches require a pre-defined taxonomy of TDM triples, they are incompatible with a realistic task definition. In short, none of the previous work is adaptable to the constantly emerging benchmarks driven by new research and innovation. In this work, we address the aforementioned problems. Unlike previous work, we (1) manually construct our dataset directly from publications to ensure complete TDMR annotations, (2) apply normalization for leaderboard construction, and (3) propose different experimental settings to simulate real-world scenarios. \medskip

Part of the scientific leaderboards can be viewed as a special type of scientific knowledge graph that includes three types of entities (Task, Dataset, Metric) and the relations between them, which have been the primary focus of the previous studies on information extraction from scientific literature [5, 7, 8, 9, 10]. Our work in the cold start scenario, in which we do not assume any pre-defined TDM triple is given, constructs such a scientific knowledge graph and links the papers to the nodes in the graph simultaneously.

\textbf{EVALUATION:} <reasoning>Contributions and position of the paper in the literature are mentioned within each paragraph rather than being summarized at the end. The first paragraph discusses previous work and their limitations, then explicitly states what the current work does differently, listing three specific contributions. The second paragraph again discusses previous studies but also explains how the current work differs by constructing a knowledge graph without assuming predefined triples and linking papers to nodes. Therefore, the evaluation score should be 1.</reasoning><score>1</score>\medskip

<END OF EXAMPLE 2> \medskip

\end{promptbox}
\caption{Prompt for Related Work Evaluation - Positioning Type (Each Paragraph) (Part 2)}
\label{fig:prompt_rw_pos_type_each_part_2}
\end{figure*}

\begin{figure*}
\centering
\begin{promptbox}[title=Prompt for Related Work Evaluation - Positioning Type (Each Paragraph) (Part 3), breakable=false]
<START OF EXAMPLE 3> \medskip

\textbf{ANSWER:} Several efforts have been made to automate the extraction and organization of performance results from scientific publications. Early work by [1] introduced a framework, TDMS-IE, for identifying task, dataset, metric, and score tuples from NLP papers, establishing a foundation for automatic leaderboard construction. Similarly, AxCell [2] proposes a robust pipeline that utilizes structural analysis and novel table segmentation techniques to extract results from machine learning papers, demonstrating significant improvements over prior methods. ORKG-Leaderboards [3] offers a systematic approach that integrates leaderboard extraction into a knowledge graph framework, enabling machine-actionable publishing and dynamic visualization of state-of-the-art performance. TELIN [4] focuses on extracting leaderboard-relevant entities from PDFs using a semi-automated approach that reduces human annotation needs through targeted entity refinement. \medskip

In addition to these extraction systems, several datasets and benchmarks have been introduced to facilitate the development and evaluation of leaderboard construction tools. LEGOBench [5] provides a large-scale benchmark derived from arXiv and PapersWithCode, and evaluates both language model-based and graph-based approaches. SciERC and its associated framework SciIE [6] support multi-task extraction of entities and relations, enabling construction of scientific knowledge graphs. SciREX [7] extends information extraction to the document level, capturing relationships that span across sections, which is critical for leaderboard generation from full papers. \medskip

Other contributions include TDMSci [8], a specialized corpus annotated with task, dataset, and metric entities, which supports the development of more accurate extraction models. SciNLP-KG [9] introduces methods for extracting entity relations from NLP literature to build a knowledge graph, highlighting its potential use in leaderboard automation. Additionally, a diachronic analysis of NLP research trends by [10] shows the evolving influence of tasks, methods, and datasets, providing a broader context for understanding the dynamic nature of scientific benchmarks.

\textbf{EVALUATION:} <reasoning>The related work section draft provided does not explicitly mention or discuss the main paper's contributions or its position. Each paragraph focuses on summarizing existing research efforts and methodologies without indicating how the current paper builds upon or differs from these works. Additionally, the final paragraph does not serve as a summary of the main paper's contributions; instead, it continues to discuss other related works without tying them back to the current study's advancements. Therefore, the evaluation score should be 0.</reasoning><score>0</score>\medskip

<END OF EXAMPLE 3> \medskip

\textbf{[ANSWER]:} \{\textit{Data instance to be evaluated}\}

\end{promptbox}
\caption{Prompt for Related Work Evaluation - Positioning Type (Each Paragraph) (Part 3)}
\label{fig:prompt_rw_pos_type_each_part_3}
\end{figure*}

\begin{figure*}
\centering
\begin{promptbox}[title=Prompt for Related Work Evaluation - Positioning Type (Final Paragraph), breakable=false]
\textbf{[QUERY]:} Your task is to write related work section that contributions and/or positioning of the paper is provided in the final paragraph as a summary.\medskip

\textbf{[CRITERIA]:} Each paragraph (except final paragraph) should focus on summarizing and comparing prior research etc., without mentioning the current paper’s contributions or its position in the literature. Then, the final paragraph at the end of the related work section should the current paper’s contributions and its positioning. The final paragraph should address aspects mentioned in previous paragraphs. Scoring rubric is as follows:\medskip

0: The answer provide contributions in each paragraph or final paragraph fails to address the points in previous paragraphs or do not mention contributions-positioning at all.\medskip

1: Only the final paragraph emphasizes the contributions-positioning by addressing the points in previous paragraphs.\medskip

\textbf{[EXAMPLES]:} \medskip

<START OF EXAMPLE 1> \medskip

\textbf{ANSWER:} \{\textit{Example answer in positioning type (Each Paragraph) (Part 1)}\}

\textbf{EVALUATION:} <reasoning>The related work draft does not mention the main paper's contributions and its position in each individual paragraph; instead, it waits until the final paragraph to outline the specific contributions of the study. The contributions are clearly summarized and detailed only in the final paragraph. Therefore, the evaluation score should be 1.</reasoning><score>1</score>\medskip

<END OF EXAMPLE 1> \medskip

<START OF EXAMPLE 2> \medskip

\textbf{ANSWER:} \{\textit{Example answer in positioning type (Each Paragraph) (Part 2)}\}

\textbf{EVALUATION:} <reasoning>Contributions and position of the paper in the literature are mentioned within each paragraph rather than being summarized at the end. The first paragraph discusses previous work and their limitations, then explicitly states what the current work does differently, listing three specific contributions. The second paragraph again discusses previous studies but also explains how the current work differs by constructing a knowledge graph without assuming predefined triples and linking papers to nodes. Therefore, the evaluation score should be 0.</reasoning><score>0</score>\medskip

<END OF EXAMPLE 2> \medskip

<START OF EXAMPLE 3> \medskip

\textbf{ANSWER:} \{\textit{Example answer in positioning type (Each Paragraph) (Part 3)}\}

\textbf{EVALUATION:} <reasoning>The related work section draft provided does not explicitly mention or discuss the main paper's contributions or its position. Each paragraph focuses on summarizing existing research efforts and methodologies without indicating how the current paper builds upon or differs from these works. Additionally, the final paragraph does not serve as a summary of the main paper's contributions; instead, it continues to discuss other related works without tying them back to the current study's advancements. Therefore, the evaluation score should be 0.</reasoning><score>0</score>\medskip

<END OF EXAMPLE 3> \medskip

\textbf{[ANSWER]:} \{\textit{Data instance to be evaluated}\}

\end{promptbox}
\caption{Prompt for Related Work Evaluation - Positioning Type (Final Paragraph)}
\label{fig:prompt_rw_pos_type_final}
\end{figure*}

\begin{figure*}
\centering
\begin{promptbox}[title=Prompt for Related Work Evaluation - Positioning Consistency (Each Paragraph) (Part 1), breakable=false]
\textbf{[QUERY]:} Your task is to write a related work section paragraph for a scientific paper that states the paper's contribution or its position among the literature.\medskip

\textbf{[CRITERIA]:} The generated paragraph should explicitly or implicitly mention the main paper's contribution or position among existing literature. Ensure that contribution and/or positioning statements should be aligned with the specific focus of the paragraph. Scoring rubric is as follows:\medskip

0: The answer paragraph fails to explicitly or implicitly mention the main paper's contribution or position among existing literature.\medskip

1: The answer paragraph explicitly or implicitly mention the main paper's contribution or position among existing literature.\medskip

\textbf{[EXAMPLES]:} \medskip

<START OF EXAMPLE 1> \medskip

\textbf{ANSWER:} In terms of data source, previous studies use either NLP-progress or paperswithcode. These sources, however, lack rigorous quality assurance, such as standardizing TDM entities across different leaderboards and ensuring complete coverage of relevant publications. Similar to our work, [1] and [2] extract TDM triples along with the results values and apply normalization for leaderboard construction. However, both studies assume a closed domain and match extracted TDM triples to a pre-defined TDM triple taxonomy. On the other hand, some studies only partially extract TDMR tuples and do not apply normalization. For example, [3] and [4] extract TDM triples without results. Therefore, these works do not deal with leaderboard construction. In addition, [6] extract the results values depending on the pre-defined TDM triples. Both [3] and [6] leverage pre-defined TDM triples in an extraction process similar to [1]. Since these approaches require a pre-defined taxonomy of TDM triples, they are incompatible with a realistic task definition. In short, none of the previous work is adaptable to the constantly emerging benchmarks driven by new research and innovation. In this work, we address the aforementioned problems. Unlike previous work, we (1) manually construct our dataset directly from publications to ensure complete TDMR annotations, (2) apply normalization for leaderboard construction, and (3) propose different experimental settings to simulate real-world scenarios. Part of the scientific leaderboards can be viewed as a special type of scientific knowledge graph that includes three types of entities (Task, Dataset, Metric) and the relations between them, which have been the primary focus of the previous studies on information extraction from scientific literature [5, 7, 8, 9, 10]. Our work in the cold start scenario, in which we do not assume any pre-defined TDM triple is given, constructs such a scientific knowledge graph and links the papers to the nodes in the graph simultaneously.

\textbf{EVALUATION:} <reasoning>The draft states the main paper's contribution and how it differs from existing literature. It outlines the limitations of previous studies and then explicitly states how the current work addresses these issues through specific contributions, such as manual dataset construction and handling cold start scenarios without pre-defined TDM triples. Therefore, the evaluation score should be 1.</reasoning><score>1</score>\medskip

<END OF EXAMPLE 1> \medskip

\end{promptbox}
\caption{Prompt for Related Work Evaluation - Positioning Consistency (Each Paragraph) (Part 1)}
\label{fig:prompt_rw_pos_cons_each_part_1}
\end{figure*}

\begin{figure*}
\centering
\begin{promptbox}[title=Prompt for Related Work Evaluation - Positioning Consistency (Each Paragraph) (Part 2), breakable=false]
<START OF EXAMPLE 2> \medskip

\textbf{ANSWER:} Several efforts have been made to automate the extraction and organization of performance results from scientific publications. Early work by [1] introduced a framework, TDMS-IE, for identifying task, dataset, metric, and score tuples from NLP papers, establishing a foundation for automatic leaderboard construction. Similarly, AxCell [2] proposes a robust pipeline that utilizes structural analysis and novel table segmentation techniques to extract results from machine learning papers, demonstrating significant improvements over prior methods. ORKG-Leaderboards [3] offers a systematic approach that integrates leaderboard extraction into a knowledge graph framework, enabling machine-actionable publishing and dynamic visualization of state-of-the-art performance. TELIN [4] focuses on extracting leaderboard-relevant entities from PDFs using a semi-automated approach that reduces human annotation needs through targeted entity refinement. In addition to these extraction systems, several datasets and benchmarks have been introduced to facilitate the development and evaluation of leaderboard construction tools. LEGOBench [5] provides a large-scale benchmark derived from arXiv and PapersWithCode, and evaluates both language model-based and graph-based approaches. SciERC and its associated framework SciIE [6] support multi-task extraction of entities and relations, enabling construction of scientific knowledge graphs. SciREX [7] extends information extraction to the document level, capturing relationships that span across sections, which is critical for leaderboard generation from full papers.

\textbf{EVALUATION:} <reasoning>The draft provides a comprehensive overview of related work but does not explicitly or implicitly mention the main paper's own contribution or how it differs from the existing literature. It talks about the efforts made but doesn't bridge into how this work builds upon them or addresses gaps they left. It focuses solely on summarizing prior work without establishing the unique position or advancement made by the current study. Therefore, the evaluation score should be 0.</reasoning>

<score>0</score>\medskip

<END OF EXAMPLE 2> \medskip

\textbf{[ANSWER]:} \{\textit{Data instance to be evaluated}\}

\end{promptbox}
\caption{Prompt for Related Work Evaluation - Positioning Consistency (Each Paragraph) (Part 2)}
\label{fig:prompt_rw_pos_cons_each_part_2}
\end{figure*}

\begin{figure*}
\centering
\begin{promptbox}[title=Prompt for Related Work Evaluation - Positioning Consistency (Final Paragraph) (Part 1), breakable=false]
\textbf{[QUERY]:} Given a context related work paragraph, your task is to write a final paragraph of related work section for a scientific paper that states the paper's contribution or its position among the literature.\medskip

\textbf{[CRITERIA]:} As a complementary to context paragraph, final paragraph should explicitly or implicitly state the current paper's contributions or its position among the literature while addressing the points from the context paragraph. In other words, contributions in the final paragraph should include the discussed points in the context paragraph. Scoring rubric is as follows:\medskip

0: Final paragraph fails to address the points in context paragraph while stating the contributions and positioning or fails to state contributions and positioning at all.\medskip

1: Final paragraph successfully address the points in context paragraph while stating the contributions and positioning of the paper.\medskip

\textbf{[EXAMPLES]:} \medskip

<START OF EXAMPLE 1> \medskip

\textbf{CONTEXT:} In a parallel research thread, scientific leaderboards have been treated as a form of scientific knowledge graphs composed of Task, Dataset, and Metric entities. Information extraction in this context has been widely studied, as seen in works such as [5, 7, 8, 9, 10]. These studies typically focus on linking scientific content to structured entities but often assume that the schema or triples are known in advance.

\textbf{ANSWER:} In contrast to prior work, our approach addresses several critical limitations. We construct our dataset manually from publications to ensure comprehensive TDMR annotations, rather than relying on incomplete or unstandardized third-party sources. We normalize TDMR tuples for accurate leaderboard construction and propose novel experimental settings that reflect realistic, dynamic scenarios, including cold-start conditions without any pre-defined TDM taxonomy. By simultaneously constructing a scientific knowledge graph and linking publications to it in an open-domain setting, our method is inherently adaptable to the evolving research landscape.

\textbf{EVALUATION:} <reasoning>The context paragraph discusses prior work on scientific leaderboards as knowledge graphs with entities like Task, Dataset, and Metric (TDM). It highlights that prior studies focus on information extraction and linking to structured entities but usually assume a pre-existing schema or set of triples. The final paragraph contrasts the main paper's contributions with this prior work by constructing a scientific knowledge graph and linking publications to it in an open-domain and adaptable manner, which directly relates to the context's theme of schema-free extraction and linking. The final paragraph clearly positions the paper among existing literature and addresses the specific limitation mentioned in the context—i.e., the assumption of known schema—by proposing methods that work without one. Therefore, the evaluation score should be 1.</reasoning><score>1</score>\medskip

<END OF EXAMPLE 1> \medskip

\end{promptbox}
\caption{Prompt for Related Work Evaluation - Positioning Consistency (Final Paragraph) (Part 1)}
\label{fig:prompt_rw_pos_cons_final_part_1}
\end{figure*}

\begin{figure*}
\centering
\begin{promptbox}[title=Prompt for Related Work Evaluation - Positioning Consistency (Final Paragraph) (Part 2), breakable=false]
<START OF EXAMPLE 2> \medskip

\textbf{CONTEXT:} In a parallel research thread, scientific leaderboards have been treated as a form of scientific knowledge graphs composed of Task, Dataset, and Metric entities. Information extraction in this context has been widely studied, as seen in works such as [5, 7, 8, 9, 10]. These studies typically focus on linking scientific content to structured entities but often assume that the schema or triples are known in advance.

\textbf{ANSWER:} Building on existing lines of research, this work explores TDMR extraction and scientific leaderboard construction. The dataset is compiled from scientific publications. The approach is designed to accommodate variations in how benchmark results are reported.

\textbf{EVALUATION:} <reasoning>The context paragraph discusses prior research treating scientific leaderboards as knowledge graphs with a focus on information extraction where the schema is often assumed to be known. The main point raised is that existing methods typically presume a known structure for the extracted information. While the final paragraph mentions the extraction of TDMR and the construction of leaderboards—aligning with the topic in the context—it does not explicitly address whether this work assumes a known schema or introduces any method for schema discovery or flexibility. Thus, the final paragraph does not clearly state how it handles the assumption of a known schema, which is a key point of the context paragraph. Therefore, the evaluation score should be 0.</reasoning><score>0</score>\medskip

<END OF EXAMPLE 2> \medskip

\textbf{[ANSWER]:} \{\textit{Data instance to be evaluated}\}

\end{promptbox}
\caption{Prompt for Related Work Evaluation - Positioning Consistency (Final Paragraph) (Part 2)}
\label{fig:prompt_rw_pos_cons_final_part_2}
\end{figure*}

\begin{figure*}
\centering
\begin{promptbox}[title=Prompt for Review Utility Evaluation - Verifiability Extraction, breakable=false]
\textbf{[QUERY]:} Your task is to write a review comment for a scientific paper. The comment should contain claims, opinions, or suggestions that require justification.\medskip

\textbf{[CRITERIA]:} A statement is considered a claim if it includes any of the following: subjective statements, suggestions or requests for changes, judgments about sections of the paper, deductions or inferred observations, and any statement where evidence or justification is required to support the claim. A statement is considered a normal (non-claim) statement if it meets these criteria: It does not contain an opinion, claim, or suggestion but consists solely of factual, descriptive content, it indicates the existence or absence of something without suggesting changes, it makes general statements about the paper that do not express an opinion, it consists of objective, factual statements that do not require verification, it includes requests for clarification or general questions, it presents logical statements or directly inferable information. Scoring rubric is as follows:\medskip

0: If the text consists solely of normal statements.\medskip

1: If the text contains claims, opinions.\medskip

\textbf{[EXAMPLES]:} \medskip

<START OF EXAMPLE 1> \medskip

\textbf{ANSWER:} You mention that dataset X is used frequently in this domain, but there is no reference supporting this statement. Could you add one?

\textbf{EVALUATION:} <reasoning>The comment points out the lack of a reference and suggests adding one but does not assert a subjective claim requiring verification. Since it is a factual observation with a request, the evaluation score should be 0.</reasoning><score>0</score>\medskip

<END OF EXAMPLE 1> \medskip

<START OF EXAMPLE 2> \medskip

\textbf{ANSWER:} Lack of novelty:- Adversarial attacks by perturbing text has been done on many NLP models and image-text models. It is nicely summarized in related work of this paper. The only new effort is to take similar ideas and apply it on video-text models.

\textbf{EVALUATION:} <reasoning>The comment claims a "lack of novelty" regarding adversarial attacks, referencing the related work section. While it provides a logical basis, additional references would strengthen it, making the claim mostly verifiable. As it is a claim, it evaluation score should be 1.</reasoning><score>1</score>\medskip

<END OF EXAMPLE 2> \medskip

\textbf{[ANSWER]:} \{\textit{Data instance to be evaluated}\}

\end{promptbox}
\caption{Prompt for Review Utility Evaluation - Verifiability Extraction}
\label{fig:prompt_ru_ver_ext}
\end{figure*}

\begin{figure*}
\centering
\begin{promptbox}[title=Prompt for Review Utility Evaluation - Verifiability (Part 1), breakable=false]
\textbf{[QUERY]:} Your task is to write a review comment for a scientific paper. The claims in your comment should be justified or proved by providing logical reasoning, using common sense, or referencing external sources.\medskip

\textbf{[CRITERIA]:} Claim justification-verification can be done either by logical reasoning supporting the claim, common sense knowledge in the field verifying the claim (e.g., referencing established practices or standards), or external references substantiating the claim. Scoring rubric is as follows:\medskip

1: The comment contains a claim without any supporting evidence or justification.\medskip

2: The comment provides some support for its claim, but the justification is vague, insufficient, or not fully articulated. Authors may struggle to follow the reasoning.\medskip

3: The comment provides support for its claim, but key elements are missing, such as specific examples, detailed explanations, or supporting references. Authors must make a significant effort to follow the justification.\medskip

4: The comment’s claim is sufficiently supported but has minor gaps. The reviewer could provide a more detailed explanation or reference.\medskip

5: The claim is thoroughly supported by explicit, sufficient, and robust evidence. This can be achieved through: - Clear and precise reasoning or explanation. - Specific and relevant references to external works or data. - Logical and unassailable common-sense arguments.\medskip

\textbf{[EXAMPLES]:} \medskip

<START OF EXAMPLE 1> \medskip

\textbf{ANSWER:} A number of claims from this paper would benefit from more in-depth analysis.

\textbf{EVALUATION:} <reasoning>The comment states that "a number of claims from this paper would benefit from more in-depth analysis." However, it does not provide any supporting reasoning or examples to justify this opinion. As a result, the comment is unverifiable, since the lack of evidence and explanation makes it difficult for the authors to understand or address the concern. Therefore, the evaluation score should be 1.</reasoning><score>1</score>\medskip

<END OF EXAMPLE 1> \medskip

<START OF EXAMPLE 2> \medskip

\textbf{ANSWER:} The relatively poor performance on nouns makes me uneasy. While I can expect TWSI to do really well due to its nature, the fact that the oracle GAP for PPDBClus is higher than most clustering approaches is disconcerting, and I would like to understand the gap better. This also directly contradicts the claim that the clustering approach is generalizable to all parts of speech (124-126), since the performance clearly isn't uniform.

\textbf{EVALUATION:} <reasoning>The comment presents a claim that the clustering approach is not generalizable to all parts of speech due to non-uniform performance, which is meant to contradict the authors' claims. However, the explanation provided is minimal, stating only that the performance is "clearly isn't uniform." While this suggests an issue, the reasoning lacks sufficient detail and examples, making it hard for the authors to fully understand the justification and address it effectively. As a result, the claim is only borderline verifiable. Therefore, evaluation score should be 2.</reasoning><score>2</score>\medskip

<END OF EXAMPLE 2> \medskip

\end{promptbox}
\caption{Prompt for Review Utility Evaluation - Verifiability (Part 1)}
\label{fig:prompt_ru_verifiability_part_1}
\end{figure*}

\begin{figure*}
\centering
\begin{promptbox}[title=Prompt for Review Utility Evaluation - Verifiability (Part 2), breakable=false]

<START OF EXAMPLE 3> \medskip

\textbf{ANSWER:} Lines 102-106 is misleading. While intersection and probs are true, "such distribution" cannot refer to the discussion in the above.

\textbf{EVALUATION:} <reasoning>The comment makes a claim that "Lines 102-106 are misleading" due to the use of "'such distribution'." While the reviewer begins to justify this by mentioning that it cannot refer to the prior discussion, the explanation is not sufficient for a full verification of the claim. More detailed reasoning or examples are needed to help the authors grasp the issue clearly, landing the comment at the evaluation score 3."</reasoning><score>3</score>\medskip

<END OF EXAMPLE 3> \medskip

<START OF EXAMPLE 4> \medskip

\textbf{ANSWER:} Also, since the dataset is artificially created, the dataset itself might have a lot of noise. For instance, the collected "pristine" set of tweets may not be pristine enough and might instead contain misinformation as well as out-of-context images. I would have liked to see more analysis around the quality of the collected dataset and the amount of noise it potentially has.

\textbf{EVALUATION:} <reasoning>The review comment makes a verifiable claim that the "dataset itself might have a lot of noise," citing the example that the 'pristine' set of tweets may include misinformation and out-of-context images. This reasoning provides a logical basis for the claim, making it mostly verifiable. However, while the argument is logically sound, it lacks detailed evidence or references, leaving minor gaps that could be addressed. Therefore, the evaluation score should be 4.</reasoning><score>4</score>\medskip

<END OF EXAMPLE 4> \medskip

<START OF EXAMPLE 5> \medskip

\textbf{ANSWER:} - line 47 - 48 "over-parametrization invariably overfits the data and results in worse performance": over-parameterization seems to be very helpful for supervised learning of deep neural networks in practice ... Also, I have seen a number of theoretical work showing the benefits of over-parametrisation e.g. [1].

\textbf{EVALUATION:} <reasoning>The review comment challenges the claim made in the paper that "over-parametrization invariably overfits the data and results in worse performance." The reviewer provides a counter-argument by stating that over-parameterization is often beneficial in supervised learning of deep neural networks and supports this with a reference to theoretical work that demonstrates these benefits. This provides a clear and precise reasoning, along with a reference, making the claim fully verifiable. Therefore, the comment is fully verifiable and should get the evaluation score 5.</reasoning><score>5</score>\medskip

<END OF EXAMPLE 5> \medskip

\textbf{[ANSWER]:} \{\textit{Data instance to be evaluated}\}

\end{promptbox}
\caption{Prompt for Review Utility Evaluation - Verifiability (Part 2)}
\label{fig:prompt_ru_verifiability_part_2}
\end{figure*}

\begin{figure*}
\centering
\begin{promptbox}[title=Prompt for Review Utility Evaluation - Actionability (Part 1), breakable=false]
\textbf{[QUERY]:} Your task is to write a review comment for a scientific paper. The comment should be actionable. Those actions should be clearly identifiable and concrete.\medskip

\textbf{[CRITERIA]:} Explicit actions or suggestions are direct or apparent. Authors can directly identify modifications they should apply to their draft. Clarification questions should be treated as explicit statements if they give a direct action. However, implicit actions need to be inferred from the comment. This includes missing parts that need to be added. Authors can deduce what needs to be done after reading the comment. For concrete actions, the authors know exactly what needs to be done and how to apply the action. However, for vague actions the authors still don’t know how to carry out this action. Scoring rubric is as follows:\medskip

1: The comment lacks meaningful information to help authors improve the paper. Authors do not know what they should do after reading the comment.\medskip

2: The comment includes an implicitly stated action or an action that can be inferred. However, the action itself is vague and lacks detail on how to apply it.\medskip

3: The comment explicitly states an action but is vague on how to execute it.\medskip

4: The comment implicitly states an action but concretely states how to implement the inferred action.\medskip

5: The comment contains an explicit action and concrete details on how to implement it. Authors know exactly how to apply it.\medskip

\textbf{[EXAMPLES]:} \medskip

<START OF EXAMPLE 1> \medskip

\textbf{ANSWER:} The hGRU architecture seems pretty ad-hoc and not very well motivated.

\textbf{EVALUATION:} <reasoning>The review comment, "The hGRU architecture seems pretty ad-hoc and not very well motivated," lacks specificity and actionable guidance for the authors. While it expresses a concern about the hGRU architecture being "ad-hoc" and "not very well motivated," it does not provide any detailed explanation or examples of why the reviewer perceives it this way. Without specific points or suggestions, the authors are left without a clear understanding of what aspects of the architecture need further clarification or improvement. hence, this comment is not actionable at all. Therefore the evaluation score should be 1.</reasoning><score>1</score>\medskip

<END OF EXAMPLE 1> \medskip

<START OF EXAMPLE 2> \medskip

\textbf{ANSWER:} The paper does not dig into the theory profs and show the convergence properties of the proposed algorithm.

\textbf{EVALUATION:} <reasoning>The comment highlights the need for the paper to delve deeper into the theory proofs and demonstrate the convergence properties of the proposed algorithm. However, it does not specify which aspects of the theory proofs require further exploration or how the authors should approach this task. The lack of concrete recommendations regarding the convergence properties leaves the authors uncertain about how to fulfill this request effectively. Due to its vague and indirect nature, this comment is barely actionable. Therefore the evaluation score should be 2.</reasoning><score>2</score>\medskip

<END OF EXAMPLE 2> \medskip

\end{promptbox}
\caption{Prompt for Review Utility Evaluation - Actionability (Part 1)}
\label{fig:prompt_ru_actionability_part_1}
\end{figure*}

\begin{figure*}
\centering
\begin{promptbox}[title=Prompt for Review Utility Evaluation - Actionability (Part 2), breakable=false]

<START OF EXAMPLE 3> \medskip

\textbf{ANSWER:} A number of claims from this paper would benefit from more in-depth analysis.

\textbf{EVALUATION:} <reasoning>The comment points out that certain claims require more in-depth analysis but does not clarify which claims need further scrutiny. As a result, the authors may not know where to focus their efforts, leading to potential misinterpretation of the feedback. Since the suggested action is direct but still lacks the necessary details for precise implementation, this comment is somewhat actionable. Therefore the evaluation score should be 3.</reasoning><score>3</score>\medskip

<END OF EXAMPLE 3> \medskip

<START OF EXAMPLE 4> \medskip

\textbf{ANSWER:} The adopted baseline models are weak. First of all, the author does not compare to Campos et al. (2020), which also uses feedback in QA tasks. Second, they also do no comparison with other domain adaptation methods, such as those work cited in Section 8.

\textbf{EVALUATION:} <reasoning>The reviewer identifies specific actions that the authors did not take, such as not comparing their work to Campos et al. (2020) and other domain adaptation methods cited in Section 8. This serves as an implicit suggestion that these comparisons should be implemented. However, rather than explicitly stating what the authors should do, the comment leaves room for interpretation. Because the required action is clear but not directly stated, the comment is mostly actionable. Therefore the evaluation score should be 4.</reasoning><score>4</score>\medskip

<END OF EXAMPLE 4> \medskip

<START OF EXAMPLE 5> \medskip

\textbf{ANSWER:} The abstract is written well and invokes intrigue early - could potentially be made even better if, for "evaluating with gold answers is inconsistent with human evaluation" - an example of the inconsistency, such as models get ranked differently is also given there.

\textbf{EVALUATION:} <reasoning>The comment explicitly states that an example of inconsistency should be provided in the abstract, specifically where it mentions "evaluating with gold answers is inconsistent with human evaluation." By directly instructing the authors to include an example, such as how models get ranked differently, it removes any uncertainty about how to proceed. Since the feedback is clear, specific, and directly actionable, the comment is fully actionable. Therefore the evaluation score should be 5.</reasoning><score>5</score>\medskip

<END OF EXAMPLE 5> \medskip

\textbf{[ANSWER]:} \{\textit{Data instance to be evaluated}\}

\end{promptbox}
\caption{Prompt for Review Utility Evaluation - Actionability (Part 2)}
\label{fig:prompt_ru_actionability_part_2}
\end{figure*}

\begin{figure*}
\centering
\begin{promptbox}[title=Prompt for Review Utility Evaluation - Grounding-specificity (Part 1), breakable=false]
\textbf{[QUERY]:} Your task is to write a review comment for a scientific paper. The comment should refer to a specific part of the paper and clearly identify the issue with that part.\medskip

\textbf{[CRITERIA]:} For fully grounded comment, the author can accurately pinpoint the section, table, figure, or unique aspect being addressed. For weak grounded comment, the author can make an educated guess but cannot precisely identify the referenced part. For specificity, the comment should detail what is wrong or missing in the referenced part. If external work is mentioned, it should also provide specific examples. Scoring rubric is as follows:\medskip

1: The comment is not grounded at all. It does not identify a specific area in the paper. The comment is highly unspecific.\medskip

2: The authors cannot confidently determine which part the comment addresses. Further, the comment does not specify what needs to be addressed in this part.\medskip

3: The authors cannot confidently determine which part the comment addresses. However, the comment clearly specifies what needs to be addressed in this part.\medskip

4: The comment explicitly mentions which part of the paper it addresses, or it should be obvious to the authors. However, this comment does not specify what needs to be addressed in this part.\medskip

5: The comment explicitly mentions which part of the paper it addresses, and it is obvious to the authors. The comment specifies what needs to be addressed in this part.\medskip

\textbf{[EXAMPLES]:} \medskip

<START OF EXAMPLE 1> \medskip

\textbf{ANSWER:} The writing should be improved. Some points in the paper is unclear to me.

\textbf{EVALUATION:} <reasoning>The comment is not grounded as it does not identify any specific part of the paper that needs improvement. It generally mentions that the writing should be improved and that some points are unclear, but it does not specify which points or sections are problematic. This lack of detail makes it difficult for the authors to know which parts of the paper to focus on for revisions. Therefore, this comment is not grounded and highly unspecific, aligning with the evaluation score 1.</reasoning><score>1</score>\medskip

<END OF EXAMPLE 1> \medskip

<START OF EXAMPLE 2> \medskip

\textbf{ANSWER:} It is always easier to show something (i.e. attention in seq2seq MTL) is not working, but the value would lie in finding out why it fails and changing the attention mechanism so that it works.

\textbf{EVALUATION:} <reasoning>The authors have some idea about which part this comment addresses, as it refers to the attention mechanism in seq2seq MTL. However, they still do not know what specifically needs to be addressed in this part. While they can guess where they mentioned that something is not working, they still cannot identify the specific issues or suggestions for improvement. The comment suggests that the value lies in understanding why the attention mechanism fails and how it can be modified to work effectively. However, without specific guidance or examples, the authors are left without clear direction. Therefore, this comment is weakly grounded and not specific, aligning with the evaluation score 2.</reasoning><score>2</score>\medskip

<END OF EXAMPLE 2> \medskip

\end{promptbox}
\caption{Prompt for Review Utility Evaluation - Grounding-specificity (Part 1)}
\label{fig:prompt_ru_grounding_part_1}
\end{figure*}

\begin{figure*}
\centering
\begin{promptbox}[title=Prompt for Review Utility Evaluation - Grounding-specificity (Part 2), breakable=false]

<START OF EXAMPLE 3> \medskip

\textbf{ANSWER:} Would the use of feature engineering help in improving the performance? Uto et al. (2020)'s system reaches a QWK of 0.801 by using a set of hand-crafted features. Perhaps using Uto et al. (2020)'s same feature set could also improve the results of this work.

\textbf{EVALUATION:} <reasoning>The comment suggests a potential improvement in the system's performance through feature engineering, specifically referencing Uto et al. (2020) as an example. While it does not explicitly mention a specific section of the paper, the authors can infer that it relates to the discussion on performance. The suggestion is specific, as it points to a particular feature set that could be considered. Therefore, the comment is weakly grounded but specific, aligning with the evaluation score 3.</reasoning><score>3</score>\medskip

<END OF EXAMPLE 3> \medskip

<START OF EXAMPLE 4> \medskip

\textbf{ANSWER:} The paper's title might not accurately reflect the central theme and main contributions of your document. A title is crucial in drawing the interest of readers and offering them a clear understanding of the primary focus of your study, guiding them from the outset.

\textbf{EVALUATION:} <reasoning>The title is explicitly mentioned, grounding the comment. Despite this, it is under-specific because it doesn’t indicate whether the problem is misleading wording, an inadequate reflection of scope, or the omission of critical terms that would better describe the paper’s content. Hence, this comment is fully grounded and under-specific. Therefore the evaluation score should be 4.</reasoning><score>4</score>\medskip

<END OF EXAMPLE 4> \medskip

<START OF EXAMPLE 5> \medskip

\textbf{ANSWER:} Why the results of Table 6 is not aligned with Table 1 (MCT-pair)? Also what about the ablation studies of MCT without the adaptive metrics.

\textbf{EVALUATION:} <reasoning>The comment is fully grounded as it explicitly mentions "Table 6" and "Table 1," allowing the authors to accurately identify the parts of the paper being addressed. Additionally, it is specific because it details the issue of alignment between the results in these tables and inquires about the ablation studies of MCT without the adaptive metrics. Therefore, this comment is categorized as fully grounded and specific, aligning with the evaluation score of 5.</reasoning><score>5</score>\medskip

<END OF EXAMPLE 5> \medskip

\textbf{[ANSWER]:} \{\textit{Data instance to be evaluated}\}

\end{promptbox}
\caption{Prompt for Review Utility Evaluation - Grounding-specificity (Part 2)}
\label{fig:prompt_ru_grounding_part_2}
\end{figure*}

\begin{figure*}
\centering
\begin{promptbox}[title=Prompt for Review Utility Evaluation - Helpfulness (Part 1), breakable=false]
\textbf{[QUERY]:} Your task is to write a review comment for a scientific paper. The comment should be useful for the authors to help improving the paper.\medskip

\textbf{[CRITERIA]:} A helpful review should be actionable, grounded on a specific part of the paper, provide justification or evidence to its claims. Scoring rubric is as follows:\medskip

1: The comment fails to identify meaningful weaknesses or suggest improvements, leaving the authors with no actionable feedback.\medskip

2: The comment identifies a weakness or improvement area but is vague, lacks clarity, or provides minimal guidance, making it only slightly beneficial for the authors.\medskip

3: The comment identifies weaknesses or areas for improvement but is incomplete or lacks depth. While the authors gain some insights, the feedback does not fully address their needs for improving the draft.\medskip

4: The comment provides clear and actionable feedback on weaknesses and areas for improvement, though it could be expanded or refined to be fully comprehensive and impactful.\medskip

5: The comment thoroughly identifies weaknesses and offers detailed, actionable, and constructive suggestions that empower the authors to significantly improve their draft.\medskip

\textbf{[EXAMPLES]:} \medskip

<START OF EXAMPLE 1> \medskip

\textbf{ANSWER:} Lack of novelty: Adversarial attacks by perturbing text has been done on many NLP models and image-text models. It is nicely summarized in related work of this paper. The only new effort is to take similar ideas and apply it on video-text models.

\textbf{EVALUATION:} <reasoning>This comment simply recaps a section already addressed in the paper and does not offer constructive guidance or specific suggestions on overcoming the stated issue of lack of novelty. Without actionable advice or an indication of how to differentiate or enhance the work, the comment is not useful to the authors and does not contribute to the paper's improvement, aligning with the evaluation score of 1: The comment is not helpful at all.</reasoning><score>1</score>\medskip

<END OF EXAMPLE 1> \medskip

<START OF EXAMPLE 2> \medskip

\textbf{ANSWER:} I don't think the probabilistic connection is drawn very well. It doesn't seem to be made formally enough to take it as anything more than motivational which is fine, but I would suggest the authors either cement this connection more formally or adjust the language to clarify.

\textbf{EVALUATION:} <reasoning>The comment highlights a potential issue with the way the probabilistic connection is presented in the paper, indicating it might lack formalism and appear purely motivational. While this could be valuable for the authors, the feedback is not specific enough to be fully actionable. The suggestion to either "cement this connection more formally" or "adjust the language to clarify" is a start, but it lacks specific guidance or examples that would make it more useful. Therefore, the authors are left with only a vague direction for improvement, making this feedback barely helpful. Therefore, the evaluation score should be 2.</reasoning><score>2</score>\medskip

<END OF EXAMPLE 2> \medskip

\end{promptbox}
\caption{Prompt for Review Utility Evaluation - Helpfulness (Part 1)}
\label{fig:prompt_ru_helpfulness_part_1}
\end{figure*}

\begin{figure*}
\centering
\begin{promptbox}[title=Prompt for Review Utility Evaluation - Helpfulness (Part 2), breakable=false]

<START OF EXAMPLE 3> \medskip

\textbf{ANSWER:} It is not clear if the proposed methodology is specific to bimanual manipulation. Just using robotic manipulation could be more appropriate.

\textbf{EVALUATION:} <reasoning>The comment identifies a potential issue with the specificity of the proposed methodology, questioning whether it is limited to bimanual manipulation or if it could be generalized to robotic manipulation. This is a relevant point that could help the authors clarify the scope and applicability of their methodology. However, the comment lacks depth and does not provide specific suggestions or guidance on how the authors might address this issue. It raises an important question but does not offer actionable steps for improvement, making it somewhat helpful. Therefore, the comment is evaluated as somewhat helpful, consistent with the evaluation score 3.</reasoning><score>3</score>\medskip

<END OF EXAMPLE 3> \medskip

<START OF EXAMPLE 4> \medskip

\textbf{ANSWER:} The authors need to perform ablation experiments to compare the proposed method with other methods (e.g., TubeR) in terms of the number of learnable parameters and GFLOPs.

\textbf{EVALUATION:} <reasoning>The review comment provides a clear and actionable suggestion for improvement by recommending the authors perform ablation experiments. It specifies the need to compare the proposed method with other methods, such as TubeR, focusing on the number of learnable parameters and GFLOPs. This feedback is direct and offers a concrete step for the authors to enhance their paper, making it mostly helpful. However, it could be further improved by explaining why these comparisons are important or how they might impact the paper's conclusions, which would elevate it to a highly helpful comment. Therefore, it aligns with a score of 4.</reasoning><score>4</score>\medskip

<END OF EXAMPLE 4> \medskip

<START OF EXAMPLE 5> \medskip

\textbf{ANSWER:} Finally, and similarly to above, i’d like to see an experiment where the image is occluded (half of the image is randomly blacked out). This (a) simulates the irregularity that is often present in neural/behavioral data (e.g. keypoint detection failed for some mice in some frames), and (b) would allow us to inspect the long-range “inference” capacity of the model, as opposed to a nearly-supervised reconstruction task. Again, these should be reasonably easy experiments to run. I’d expect to see all of these experiments included in a final version (unless the authors can convince me otherwise).

\textbf{EVALUATION:} <reasoning>The comment is highly helpful as it provides a clear and specific suggestion for an additional experiment that could significantly enhance the paper. By proposing an experiment where the image is occluded, the reviewer not only identifies a potential area for improvement but also explains the rationale behind it—simulating irregularities in data and testing the model's inference capacity. This feedback is actionable and constructive, offering the authors a concrete way to strengthen their work. The expectation that these experiments be included unless justified otherwise further emphasizes the importance of this suggestion, making it a valuable contribution to the paper's development.</reasoning><score>5</score>\medskip

<END OF EXAMPLE 5> \medskip

\textbf{[ANSWER]:} \{\textit{Data instance to be evaluated}\}

\end{promptbox}
\caption{Prompt for Review Utility Evaluation - Helpfulness (Part 2)}
\label{fig:prompt_ru_helpfulness_part_2}
\end{figure*}

\begin{figure*}
\centering
\begin{promptbox}[title=Prompt for Novelty Alignment Evaluation (Part 1), breakable=false]
\textbf{[QUERY]:} Your task is to write two assessments regarding novelty of a scientific paper. Each one should lead to implicitly/explicitly a verdict of either "novel" or "not novel". The assessments should be aligned in terms if their novelty decision. \medskip

\textbf{[CRITERIA]:} Assessments with verdict "novel" states that the paper introduces new, original, or significant ideas, methods, metrics, or frameworks; emphasizes a meaningful, notable, or important contribution; describes the work as advancing the field in a substantive or distinct way. Even if it is not groundbreaking fundamental new contribution, the paper can be still novel based on its significant contributions. On the other hand, assessments implying "not novel" state that the contribution is totally incremental, limited, weak, already known; emphasize that prior work already covers most of the ideas or findings; say the paper lacks significant, substantial, or original contributions; describe the work as mainly empirical, confirmatory, or replicating existing knowledge without new insights. You need to check whether these two assessments' final conclusion is the same or not. It is possible that two assessments can have similar observations on some points but they can arrive different conclusions. Your evaluation should be according to final conclusions. Evaluation two assessments should be binary. Scoring rubric is as follows.\medskip

0: Two assessments do not come to the same conclusion in terms of novelty verdict.\medskip

1: Two assessments come to the same conclusion in terms of novelty verdict.\medskip

\textbf{[EXAMPLES]:} \medskip

<START OF EXAMPLE 1> \medskip

\textbf{ANSWER:} \textbf{Assessment 1:} This paper introduces a novel framework for analyzing the self-improvement capabilities of large language models, centering on the proposal of the generation-verification gap (GV-gap) as a new metric to quantify the limits of self-improvement. The GV-gap is clearly defined and illustrated with real-world examples, offering a fresh perspective for measuring and understanding where self-improvement may be fundamentally constrained. This approach represents a meaningful contribution to the research direction, as it provides the community with a new tool for future studies on model self-improvement. While the concept of self-improvement in language models has been explored in prior work, the introduction of the GV-gap offers a distinct and unified way to formalize and assess these capabilities. The novelty of the paper lies in this new quantification method, which can help clarify and advance discussions in the field. However, the practical application of the GV-gap is somewhat limited by the noisiness of real-world utility functions, as acknowledged by the authors, which may affect the robustness of the measurements. Despite this limitation, the proposed framework and metric are likely to be valuable for future research, marking the work as a notable and useful contribution to ongoing efforts in understanding and improving language model self-improvement. \textbf{Assessment 2:} This submission offers a comprehensive and systematic empirical/theoretical study of LLM self-improvement, with its main novelty being the formalization and central use of the "generation-verification gap" (GV-Gap) as a unifying metric. While the concept of a gap between generation and verification is present in prior work, the explicit metric and its application across models and tasks, as well as the discovery of a scaling law for GV-Gap, are new contributions. The paper also provides a detailed, cross-model analysis of verification mechanisms, including ensemble verification, which is a substantive but incremental extension of existing meta-judging and reward aggregation methods. However, the submission tends to overstate the lack of systematic analysis and diversity in prior work, and does not fully engage with risks such as bias and diversity collapse highlighted in the literature. Overall, the work represents a significant incremental advance in empirical rigor and formalization, rather than a fundamentally new paradigm for LLM self-improvement.

\end{promptbox}
\caption{Prompt for Novelty Alignment Evaluation (Part 1)}
\label{fig:prompt_novelty_part_1}
\end{figure*}

\begin{figure*}
\centering
\begin{promptbox}[title=Prompt for Novelty Alignment Evaluation (Part 2), breakable=false]
\textbf{EVALUATION:} <reasoning>Assessment 1 clearly concludes that the work is novel, emphasizing that the GV-gap constitutes a "meaningful contribution" and "new metric," marking the paper as a "notable and useful contribution." Assessment 2, while calling the work "a significant incremental advance," still attributes new contributions (formalizing GV-Gap, discovering a scaling law). It does not declare the work non-novel or merely confirmatory; it frames the contribution as incremental "but still novel". Thus both assessments align on a "novel" verdict, the score should be 1.</reasoning><score>1</score>\medskip

<END OF EXAMPLE 1> \medskip

<START OF EXAMPLE 2> \medskip

\textbf{ANSWER:} \textbf{Assessment 1:} This paper presents a comprehensive experimental analysis of self-improvement in Large Language Models (LLMs), focusing on the concept of the generation-verification gap and its relationship to model pre-training computational effort. While the study offers a modular framework and conducts extensive experiments to examine scaling phenomena and conditions for self-improvement, its novelty is limited. Most of the conclusions, such as the monotonic scaling of the verification gap with pre-training FLOPs and the identification of saturation limits, are already established in the literature. As a result, the paper does not provide fundamentally new insights or advances beyond what is already known. The contribution is primarily empirical, and while the analysis is thorough, it does not introduce novel theoretical perspectives or experimental findings that significantly advance the field. Consequently, the work’s impact on the community is relatively weak from a novelty standpoint. \textbf{Assessment 2:} This submission offers a comprehensive and systematic empirical/theoretical study of LLM self-improvement, with its main novelty being the formalization and central use of the "generation-verification gap" (GV-Gap) as a unifying metric. While the concept of a gap between generation and verification is present in prior work, the explicit metric and its application across models and tasks, as well as the discovery of a scaling law for GV-Gap, are new contributions. The paper also provides a detailed, cross-model analysis of verification mechanisms, including ensemble verification, which is a substantive but incremental extension of existing meta-judging and reward aggregation methods. However, the submission tends to overstate the lack of systematic analysis and diversity in prior work, and does not fully engage with risks such as bias and diversity collapse highlighted in the literature. Overall, the work represents a significant incremental advance in empirical rigor and formalization, rather than a fundamentally new paradigm for LLM self-improvement.

\textbf{EVALUATION:} <reasoning>Assessment 1 concludes the paper is not novel, emphasizing that it offers only empirical, already-known findings, and no fundamentally new insights. Assessment 2 acknowledges the work as a **significant incremental advance** with *new contributions* (formalizing the metric, discovering a scaling law). Although incremental, it is still treated as novel. Since assessments do not align in terms of novelty of the paper, the evaluation score should 0.</reasoning><score>0</score>\medskip

<END OF EXAMPLE 2> \medskip

\textbf{[ANSWER]:} \textbf{Assessment 1:} \{\textit{First novelty assessment}\} \textbf{Assessment 2:} \{\textit{Second novelty assessment}\}

\end{promptbox}
\caption{Prompt for Novelty Alignment Evaluation (Part 2)}
\label{fig:prompt_novelty_part_2}
\end{figure*}

\begin{figure*}
\centering
\begin{promptbox}[title=Prompt for Revision Evaluation - Relatedness (Part 1), breakable=false]
\textbf{[QUERY]:} Your task is to revise a scientific text according to given an instruction. The revised text should address the points in the instruction.\medskip

\textbf{[CRITERIA]:} The generated revision should correctly address the instruction. Scoring rubric is as follows.\medskip

0: The model revision does not address the instruction.\medskip

1: The revision follows the requirement of the instruction.\medskip

\textbf{[EXAMPLES]:} \medskip

<START OF EXAMPLE 1> \medskip

\textbf{ORIGINAL TEXT:} MNIST dataset with 1000 training images. All external regularization schemes except learning rate decay and batch normalization have been turned off. We perform the following experiments : 1 ) Full-batch gradient descent with $\beta$ = (i.e., GD) for various learning rate h and the best test accuracy is noted (in Figure 2) to be 95 . Full-batch gradient descent with momentum (GD+M) performed for various $\beta$ with a fixed step-size h - 0 . 1 and the best test-accuracy is noted (in Figure 3) to be 96 . Our observation is that the best performance of GD (across all learning rates) is worse than the best performance of (GD+M) (across all $\beta$ ’s). (Cohen et al., 2021) showed that gradient descent (GD) has an overwhelming tendency to increase the sharpness 2 till it reaches 2 h , called “the edge of stability”. And for (GD+M), the sharpness can reach up to 2(1+ $\beta$ ) h , hence allowing it to enter a sharper region before becoming unstable. As greater allowable sharpness for (GD+M) than that of (GD) may suggest a higher test accuracy for (GD), this is not what we observe from the above experiment. We think the  implicit regularization  for (GD+M) plays a part in it. We believe IGR for momentum outweighs the sharpness effect in achieving better test accuracy.

\textbf{INSTRUCTION:} Summarize the second half of the paragraph to make the paragraph shorter.

\textbf{ANSWER:} MNIST dataset with 1000 training images. All external regularization schemes except learning rate decay and batch normalization have been turned off. We perform the following experiments : 1 ) Full-batch gradient descent with  = (i.e., GD) for various learning rates and the best test accuracy is noted to be 95. For (GD+M) performed for various  with a fixed step-size h  0. 1 and the best test-accuracy is noted to be 96. Our observation is that the best performance of GD (across all learning rates) is worse than the best performance of (GD+M) across all ’s).

\textbf{EVALUATION:} <reasoning>The provided answer omits the entire theoretical discussion rather than summarizing it, so it does not properly address the instruction. Therefore the evaluation score should be 0.</reasoning><score>0</score>\medskip

<END OF EXAMPLE 1> \medskip

\end{promptbox}
\caption{Prompt for Revision Evaluation - Relatedness (Part 1)}
\label{fig:prompt_revision_rel_part_1}
\end{figure*}

\begin{figure*}
\centering
\begin{promptbox}[title=Prompt for Revision Evaluation - Relatedness (Part 2), breakable=false]

<START OF EXAMPLE 2> \medskip

\textbf{ORIGINAL TEXT:} MNIST dataset with 1000 training images. All external regularization schemes except learning rate decay and batch normalization have been turned off. We perform the following experiments : 1 ) Full-batch gradient descent with $\beta$ = (i.e., GD) for various learning rate h and the best test accuracy is noted (in Figure 2) to be 95 . Full-batch gradient descent with momentum (GD+M) performed for various $\beta$ with a fixed step-size h - 0 . 1 and the best test-accuracy is noted (in Figure 3) to be 96 . Our observation is that the best performance of GD (across all learning rates) is worse than the best performance of (GD+M) (across all $\beta$ ’s). (Cohen et al., 2021) showed that gradient descent (GD) has an overwhelming tendency to increase the sharpness 2 till it reaches 2 h , called “the edge of stability”. And for (GD+M), the sharpness can reach up to 2(1+ $\beta$ ) h , hence allowing it to enter a sharper region before becoming unstable. As greater allowable sharpness for (GD+M) than that of (GD) may suggest a higher test accuracy for (GD), this is not what we observe from the above experiment. We think the  implicit regularization  for (GD+M) plays a part in it. We believe IGR for momentum outweighs the sharpness effect in achieving better test accuracy.

\textbf{INSTRUCTION:} Summarize the second half of the paragraph to make the paragraph shorter.

\textbf{ANSWER:} MNIST dataset with 1000 training images. All external regularization schemes except learning rate decay and batch normalization have been turned off. We perform experiments with full-batch gradient descent (GD) and full-batch gradient descent with momentum (GD+M) for various learning rates and $\beta$ values. Our results show that GD+M outperforms GD, achieving a best test accuracy of 96 compared to 95 for GD. This contradicts the expectation that GD+M's higher allowable sharpness would lead to better test accuracy. We attribute this to the implicit regularization effect of momentum, which outweighs the sharpness effect in achieving better test accuracy.

\textbf{EVALUATION:} <reasoning>The revision clearly summarizes the theoretical discussion in a shorter form while preserving the key ideas: GD+M’s higher accuracy, the contradiction with sharpness-based expectations, and the role of implicit regularization. Because it successfully condenses the second half of the paragraph as instructed, it meets the requirement. Therefore the evaluation score should be 1.</reasoning><score>1</score>\medskip

<END OF EXAMPLE 2> \medskip

\textbf{[ORIGINAL TEXT]:} \{Original text to be revised\}

\textbf{[INSTRUCTION]:} \{Revision instruction\}

\textbf{[ANSWER]:} \{{Revision to be evaluated}\}

\end{promptbox}
\caption{Prompt for Revision Evaluation - Relatedness (Part 2)}
\label{fig:prompt_revision_rel_part_2}
\end{figure*}

\begin{figure*}
\centering
\begin{promptbox}[title=Prompt for Revision Evaluation - Correctness (Part 1), breakable=false]
\textbf{[QUERY]:} Your task is to revise a scientific text according to given an instruction. The revised text should be in a better state after applying the instructions.\medskip

\textbf{[CRITERIA]:} The revised text should be a good quality revision proposition that should replace the original paragraph in the scientific article. Scoring rubric is as follows.\medskip

0: The revision is not a better version that can replace the original paragraph in the scientific article.\medskip

1: The revision is a better version that should replace the original paragraph in the scientific article.\medskip

\textbf{[EXAMPLES]:} \medskip

<START OF EXAMPLE 1> \medskip

\textbf{ORIGINAL TEXT:} SuperWeights, which are linear combination of templates which get reused by multiple layers . These SuperWeights capture a single operation on the input features ( e.g ., edge or texture detectors), and are themselves generated via a weighted combination of one or more templates of trainable parameters held by Weight Templates. Thus, to generate the weights for a single layer, we must first construct SuperWeights from the trainable parameters held by Weight Templates (discussed in Section 3.1), and then concatenate together all SuperWeights used by the layer to create its final weights (process illustrated in center-right column of Figure 2).

\textbf{INSTRUCTION:} Make the first half of the paragraph shorter by merging the two sentences and removing the details about how superweights are generated.

\textbf{ANSWER:} SuperWeights, linear combinations of templates reused by multiple layers, capture a single operation on input features (e.g., edge or texture detectors). Generated via weighted combinations of trainable parameters held by Weight Templates, they are then concatenated to create a layer's final weights.

\textbf{EVALUATION:} <reasoning>The instruction asks to make the first half shorter by merging the two sentences and *removing the details about how SuperWeights are generated*. The provided answer does merge the sentences and shortens the text, but it still includes the generation detail (“Generated via weighted combinations of trainable parameters held by Weight Templates”), which the instruction explicitly requests to remove. Because the revision does not fully follow the instruction and retains disallowed details, it is not an improved replacement for the scientific article. Therefore, the evaluation score should be 0.</reasoning><score>0</score>\medskip

<END OF EXAMPLE 1> \medskip

\end{promptbox}
\caption{Prompt for Revision Evaluation - Correctness (Part 1)}
\label{fig:prompt_revision_cor_part_1}
\end{figure*}

\begin{figure*}
\centering
\begin{promptbox}[title=Prompt for Revision Evaluation - Correctness (Part 2), breakable=false]

<START OF EXAMPLE 2> \medskip

\textbf{ORIGINAL TEXT:} SuperWeights, which are linear combination of templates which get reused by multiple layers . These SuperWeights capture a single operation on the input features ( e.g ., edge or texture detectors), and are themselves generated via a weighted combination of one or more templates of trainable parameters held by Weight Templates. Thus, to generate the weights for a single layer, we must first construct SuperWeights from the trainable parameters held by Weight Templates (discussed in Section 3.1), and then concatenate together all SuperWeights used by the layer to create its final weights (process illustrated in center-right column of Figure 2).

\textbf{INSTRUCTION:} Make the first half of the paragraph shorter by merging the two sentences and removing the details about how superweights are generated.

\textbf{ANSWER:} SuperWeights, which are linear combinations of templates, capture a single operation on the input features (e.g., edge or texture detectors). Thus, to generate the weights for a single layer, we must first construct SuperWeights from the trainable parameters held by Weight Templates (discussed in Section 3.1), and then concatenate together all SuperWeights used by the layer to create its final weights (process illustrated in center-right column of Figure 2).

\textbf{EVALUATION:} <reasoning>The instruction requires merging the first two sentences and removing details about how SuperWeights are generated. The revision successfully merges the sentences into one concise sentence describing what SuperWeights are and what they capture. It also removes the explicit description of their generation process. The remainder of the paragraph is kept intact and coherent. Overall, the revision is cleaner, follows the instruction, and is suitable to replace the original scientific text. Therefore, the evaluation score should be 1.</reasoning><score>1</score>\medskip

<END OF EXAMPLE 2> \medskip

\textbf{[ORIGINAL TEXT]:} \{Original text to be revised\}

\textbf{[INSTRUCTION]:} \{Revision instruction\}

\textbf{[ANSWER]:} \{{Revision to be evaluated}\}

\end{promptbox}
\caption{Prompt for Revision Evaluation - Correctness (Part 2)}
\label{fig:prompt_revision_cor_part_2}
\end{figure*}

\end{document}